\documentclass{article}
\usepackage[utf8]{inputenc}
\usepackage[T1]{fontenc}
\usepackage{lmodern}
\usepackage[margin=2.5cm]{geometry}
\usepackage{microtype}

\usepackage{mathtools}   
\usepackage{amssymb, amsthm, amsfonts, mathrsfs, bm, bbm}

\usepackage{authblk}
\usepackage[numbers,square]{natbib}

\usepackage{graphicx, xcolor, float, multirow, makecell, booktabs}
\usepackage{caption, subcaption}
\usepackage{algorithm, algpseudocode}
\usepackage{tikz}
\usetikzlibrary{arrows.meta, positioning, decorations.pathreplacing}
\usepackage{indentfirst}
\usepackage{tocloft}
\usepackage{dsfont}
\usepackage[colorlinks=true, linkcolor=blue!60!black, citecolor=blue!60!black, urlcolor=blue!60!black]{hyperref}

\title{\textbf{SOTS\\[2pt]
Signature-based Optimal Transport Selection}\\[6pt]
{\large\bfseries Coreset Selection for Irregular Multivariate Time Series}}

\author[1,2]{Khalid Akhmouch\thanks{Corresponding author: \texttt{khalid.akhmouch@polytechnique.edu}}}
\author[2]{Eric Jauer}
\affil[1]{École Polytechnique, Institut Polytechnique de Paris, Palaiseau, France}
\affil[2]{Framatome, Paris La Défense, France}

\date{}

\begin{document}
\maketitle
\footnotetext{Code: \url{https://github.com/khalidchdid/SOTS}}

\begin{abstract}
Training data selection for machine learning models is a critical bottleneck in industrial applications where expert annotation is costly and time-consuming. Existing coreset selection methods for time series either ignore temporal structure entirely or cannot handle variable-length, irregularly sampled, multi-channel signals, the setting most common in industrial monitoring and testing. We introduce SOTS (Signature-based Optimal Transport Selection), an unsupervised coreset selection framework for irregular multivariate time series (IMTS) classification. Each sample is mapped to a truncated path signature, capturing the joint geometric structure of its channels, and the coreset is selected by greedily minimizing the Sliced Wasserstein distance between the empirical signature distributions of the candidate subset and the full dataset, a criterion for which we establish a convergence guarantee for random coresets. We validate SOTS on a real-world, highly imbalanced industrial I\&C testing dataset, where SOTS-Greedy exceeds 95\% of full-dataset accuracy using only 30\% of the training data, outperforming standard coreset baselines at most budgets. This advantage generalizes to public UEA regular multivariate time series benchmarks specifically when class-discriminative information is purely geometric.
\\
Code is available at \url{https://github.com/khalidchdid/SOTS}.
\end{abstract}


\newtheorem{example}{Example}
\newtheorem{theorem}{Theorem}
\newtheorem{proposition}{Proposition}
\newtheorem{lemma}{Lemma}
\newtheorem{corollary}{Corollary}
\newtheorem{definition}{Definition}
\newtheorem{remark}{Remark}
\newtheorem{fact}{Fact}

\theoremstyle{remark}

\definecolor{primary}{RGB}{31, 56, 100}
\definecolor{secondary}{RGB}{46, 117, 182}
\definecolor{accent}{RGB}{237, 125, 49}
\definecolor{darkgray}{RGB}{80, 80, 80}

\newcommand{\subtitle}[1]{%
  \posttitle{%
    \par\large#1\par\vskip0.5em
  }
}

\newcommand{\shuffle}{\mathbin{\sqcup\mkern-3.3mu\sqcup}}

\newpage

\tableofcontents

\newpage

\section{Introduction}

Not every machine learning problem is a big-data problem. Large language
models are trained on trillions of tokens with correspondingly large
compute budgets, but many industrial, medical, and financial applications
rely on task-specific models trained on far smaller datasets, and often
with modest compute. When data is scarce or expensive to label, the
question shifts from how much data to collect to which data is worth
collecting: not every point in a dataset contributes equally to a
downstream task, and identifying the informative ones from the redundant
or uninformative ones becomes central rather than incidental. Simple
filtering heuristics, removing exact duplicates or points that clearly
fall outside the task's scope, rarely shrink a dataset by much, precisely
because they do not reason about task relevance. Conversely, discarding
points that are inconvenient, for instance those with missing values that
are hard to impute, can remove information the task actually depends on.
Formalizing what makes a subset of data \emph{representative} of a
learning task, rather than simply small, is the coreset selection
problem. In this paper we study coreset selection for classification: we
assume a downstream classifier is fixed, but do not yet know which
instances to feed it for best performance. This setting is particularly
relevant when data arrives unlabeled: selecting the most informative
instances \emph{before} labeling, rather than after, directly reduces the
labeling effort required. In our motivating industrial setting, for
example, a single test report can flag tens of thousands of discrepancies
for a verification engineer to manually review; selecting a few hundred
of the most representative ones for labeling, rather than the full
report, would save a considerable amount of time and ensure better labeling of data.

We focus specifically on irregular multivariate time series (IMTS), the
most common form of time-indexed data in the industrial settings that
motivate this work. Multivariate time series arise whenever several
sensors jointly measure a system over time, and are ubiquitous across
domains, from clinical monitoring to industrial process control. In many
industrial contexts, however, the channels of a multivariate series are
not recorded synchronously: sensors sample at different rates, drop out
for periods of time, or are only active over different intervals, so that
no single, shared timestamp grid exists across channels. This is not an
edge case; it is close to the default in real deployments. In our own
application, comparing an oracle and a simulated signal to detect false
positives in nuclear Instrumentation and Control (I\&C) software testing
\citep{anas}, the two compared signals are independently
sampled and routinely differ in rate, producing exactly this kind of
irregularity.

The literature on IMTS classification is comparatively limited, and
coreset selection specifically for IMTS is scarcer still. On the
classification side, recent architectures such as GRU-D
\citep{grud}, SEFT \citep{seft}, and adapted
Transformer variants \citep{attention,seft} handle
irregularity by encoding missingness directly, typically via masking
indicators or continuous time embeddings, rather than by imputing to a
regular grid first. On the coreset side, the closest prior work we are
aware of is \cite{clustering}, who construct coresets with
provable size guarantees for time series \emph{clustering} under a
parametric Gaussian mixture model with autocorrelation. Their setting
differs from ours in three respects: it targets clustering rather than
classification, it assumes regularly time-indexed panel data rather than
irregular, asynchronously sampled channels, and its notion of similarity
is defined by the parametric generative model rather than by any
geometric property of the observed trajectories. Closer in spirit to our
approach, \cite{pallage} use the sliced Wasserstein distance
directly for unsupervised training data selection in an industrial
setting, filtering points whose removal would most reduce the empirical
distributional distance to the rest of the dataset. Their method targets
outlier filtering on raw, fixed-dimensional features, however, and does
not address either the irregularity of multivariate time series or a
representation capturing their geometric structure; the underlying
intuition, that a sliced-Wasserstein criterion can guide data selection
in industrial settings, is one we share, but apply to a different problem, through a different representation and with a different algorithm.

Coreset selection for classification is also, more fundamentally,
under-specified without reference to what the coreset is meant to
approximate. A coreset is conventionally defined with respect to a fixed
dataset and a cost function: the coreset is good if a learning algorithm
achieves similar cost on it as on the full dataset. This framing,
however, treats the dataset as an end in itself, when in practice a
dataset is only ever a finite sample from an underlying data distribution,
and the quantity a learning algorithm actually seeks to minimize is
expected loss under that distribution, not loss on the particular sample
observed. \cite{claici} make this point explicitly,
proposing measure coresets that are evaluated against the underlying
distribution rather than the sample, and showing that a natural way to
construct such coresets is to directly minimize a Wasserstein distance
between the coreset's empirical distribution and the target measure, an
idea with roots in the broader coreset literature
\citep{agarwal,coresetliterature}. Their construction
targets continuous measures in low-dimensional Euclidean space via
stochastic gradient descent and full Wasserstein distances, and is
demonstrated on clustering, SVM classification, and Bayesian inference;
it does not address sequential or time-indexed data, and offers no
mechanism for handling irregularity or extracting geometric structure
from a multivariate trajectory. We adopt the same underlying principle,
that a coreset should approximate the full dataset's distribution rather
than merely perform well on it, but instantiate it for IMTS classification
through a representation and a selection procedure tailored to that
setting.

To address these gaps, we propose SOTS, a coreset selection method for
IMTS classification tailored to tasks whose labels are determined by the
\emph{geometric} structure of a sample: the shape traced jointly by its
channels, rather than the value of any single channel considered in
isolation. Concretely, geometric information refers to what a human
expert would visually attend to when labeling a sample, for instance,
the area enclosed between two compared signals, or the offset between
their trajectories, considered jointly rather than pointwise. To capture
this, we map each IMTS instance into its truncated path signature, a
representation from rough path theory shown to characterize, up to equivalence, a path's
geometry
\citep{lyons1998differential,cuchiero} and to perform strongly on
time series classification benchmarks \citep{morrill}.
Because the signature is defined per-channel and does not require a
shared timestamp grid, it is a natural fit for irregularly sampled
multivariate series. Given this representation, we then select a coreset
by greedily minimizing the sliced Wasserstein distance
\citep{w-sw,generalizedsw} between the
coreset's empirical distribution and the full dataset's, in signature
space, a computationally tractable surrogate for the full Wasserstein
distance that admits both a convergence guarantee for random coresets
(Proposition~\ref{prop:convergence}) and, as we show empirically, a close
practical coupling with downstream classification performance,
particularly in the low-budget regime where coreset selection is most
useful.

This design targets a specific, testable hypothesis: SOTS should be most
effective when class-discriminative information is itself geometric, and
offer comparatively little advantage otherwise. We validate this
directly on our motivating industrial application, an irregularly
sampled, class-imbalanced I\&C testing dataset, and additionally test it
outside this regime on three datasets from the UEA multivariate time
series archive \citep{uea}, chosen to span both geometric and
non-geometric discriminative signals under regular sampling.

\paragraph{Our contributions.}
In this work, we introduce SOTS, a novel coreset selection method for irregular multivariate
time series (IMTS) classification that combines a truncated path
signature representation, capturing the joint geometric structure of a
sample's channels, with greedy minimization of the sliced Wasserstein
distance between the coreset's and the full dataset's empirical
distributions. This combination yields an unsupervised, label-agnostic
selection criterion that requires no synchronized timestamp grid across
channels, and is efficient to compute at scale via a GPU-parallelized
greedy procedure (Algorithm~\ref{alg:greedy}). Our contributions are
as follows:

\begin{enumerate}
\item we establish a convergence guarantee for random coresets in
signature space, obtained by combining empirical-measure convergence
rates in Wasserstein distance~\citep{fournier}, the
H\"older-type equivalence between sliced Wasserstein and Wasserstein
distances~\citep{w-sw}, and structural properties
of truncated path signatures (Section~\ref{subsec:signatures});
\item we empirically validate the link between this distributional
distance and downstream classification performance, showing that sliced
Wasserstein distance decreases and classification accuracy increases as
coreset budget grows, direct empirical support for
minimizing this distance as a coreset selection criterion
(Figure~\ref{fig:sw-perf});
\item we validate SOTS empirically on a real-world, irregularly
sampled, class-imbalanced industrial I\&C testing
dataset (Section~\ref{subsec:ind-setup}), where it wins outright in $50\%$ of
all (budget, metric) configurations and ranks among the top two methods
in $82\%$, substantially outperforming established coreset selection
baselines; we further test generalization on three datasets from the
UEA multivariate time series archive (Section~\ref{subsec:uea}) under
regular sampling, and find that SOTS's advantage is strongest when class-discriminative information is
itself geometric.
\end{enumerate}

\newpage

\section{Methods}
\label{sec:methods}

We present \textsc{SOTS} (\textbf{S}ignature-based \textbf{O}ptimal
\textbf{T}ransport \textbf{S}election), an unsupervised
coreset selection framework for irregular multivariate time series
(IMTS). The method proceeds in three stages: (i) each time series is
encoded as a fixed-dimensional vector through a truncated path
signature; (ii) the discrepancy between a candidate coreset and the
full dataset is measured by the Sliced Wasserstein distance between
the empirical distributions of these signature vectors; and (iii) the
coreset is constructed by greedily minimizing this distance. We first
formalize the problem and present the framework in its general form
for arbitrary IMTS (Sections~\ref{subsec:problem}
to~\ref{subsec:algorithms}), then detail its instantiation on the
industrial monitoring data that motivated this work
(Section~\ref{subsec:ind-inst}).

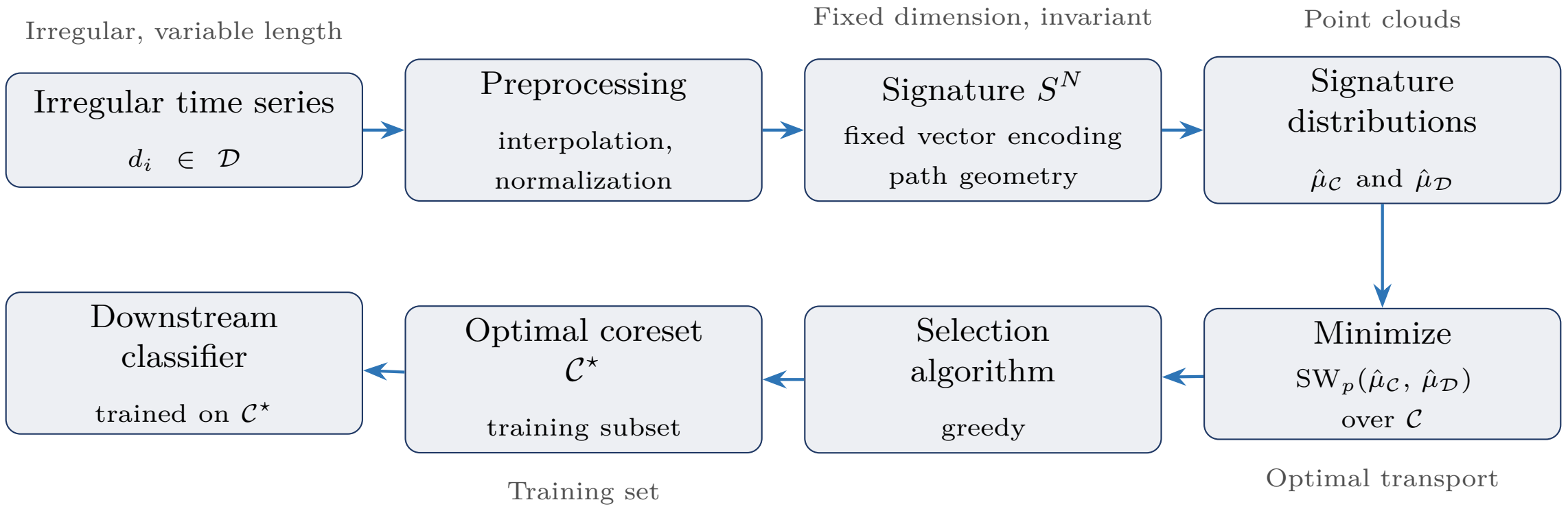
\begin{figure}[H]
\centering

\begin{tikzpicture}[
    node distance=0.4cm and 0.6cm,
    box/.style={rectangle, rounded corners=4pt, draw=primary, fill=primary!8,
                text width=3.2cm, align=center, minimum height=1.1cm, font=\small},
    arrow/.style={-{Stealth}, thick, secondary},
    arrowdown/.style={-{Stealth}, thick, secondary},
    label/.style={font=\scriptsize, darkgray, align=center}
]
    \node[box] (raw) {Irregular time series\\[4pt]
      \scriptsize $d_i \in \mathcal{D}$};
    \node[box, right=0.4cm of raw] (pre) {Preprocessing\\[4pt]
        \scriptsize interpolation, normalization};
    \node[box, right=0.4cm of pre] (sig) {Signature $S^N$\\[4pt]
      \scriptsize fixed vector encoding\\path geometry};
    \node[box, right=0.4cm of sig] (dist) {Signature\\distributions\\[4pt]
      \scriptsize $\hat{\mu}_{\mathcal{C}}$ and $\hat{\mu}_{\mathcal{D}}$};
    \draw[arrow] (raw) -- (pre);
    \draw[arrow] (pre) -- (sig);
    \draw[arrow] (sig) -- (dist);

    \node[box, below=1.0cm of raw] (train) {Downstream\\classifier\\[4pt]
          \scriptsize trained on $\mathcal{C}^\star$};
    \node[box, below=1.0cm of pre] (core) {Optimal coreset\\$\mathcal{C}^\star$\\[4pt]
      \scriptsize training subset};
    \node[box, below=1.0cm of sig] (algo) {Selection\\algorithm\\[4pt]
      \scriptsize greedy};
    \node[box, below=1.0cm of dist] (ot) {Minimize\\[4pt]
      \scriptsize $\mathrm{SW}_p(\hat{\mu}_{\mathcal{C}},\, \hat{\mu}_{\mathcal{D}})$\\over $\mathcal{C}$};

    \draw[arrow] (core) -- (train);
    \draw[arrow] (algo) -- (core);
    \draw[arrow] (ot) -- (algo);
    \draw[arrowdown] (dist) -- (ot);

    \node[above=0.15cm of raw, label] {Irregular, variable length};
    \node[above=0.15cm of sig, label] {Fixed dimension, invariant};
    \node[above=0.15cm of dist, label] {Point clouds};
    \node[below=0.15cm of core, label] {Training set};
    \node[below=0.15cm of ot, label] {Optimal transport};
\end{tikzpicture}

\caption{Overview of the SOTS pipeline. Each irregular time
series $d_i \in \mathcal{D}$ is mapped to a fixed-dimensional
signature vector $\Phi(d_i)$ (Section~\ref{subsec:signatures}),
turning the dataset and any candidate coreset into empirical
measures $\hat{\mu}_{\mathcal{D}}$ and $\hat{\mu}_{\mathcal{C}}$
on $\mathbb{R}^{d}$. The coreset $\mathcal{C}^{\star}$ is
selected by minimizing the Sliced Wasserstein distance between
the two measures (Section~\ref{subsec:sw}), greedily
(Section~\ref{subsec:algorithms}), and used to train the
downstream classifier.}
\label{fig:pipeline}
\end{figure}

\subsection{Problem Formalization}
\label{subsec:problem}

\begin{figure}[h]
\centering
\begin{tikzpicture}[scale=1.0]

\def\tAa{0.0}
\def\tAb{1.3}
\def\tAc{2.6}
\def\tAd{3.4}
\def\tAe{4.7}
\def\tAf{6.0}

\def\tBa{0.3}
\def\tBb{0.9}
\def\tBc{2.0}
\def\tBd{3.9}
\def\tBe{5.2}
\def\tBf{5.9}

\def\tCa{0.5}
\def\tCb{2.2}
\def\tCc{2.9}
\def\tCd{4.1}
\def\tCe{4.6}
\def\tCf{6.0}

\def\rowA{5.0}
\def\rowB{2.8}
\def\rowC{0.6}

\draw[-{Stealth}, thick] (-0.6,-0.4) -- (6.8,-0.4) node[right] {$t$};

\node[left, font=\small] at (-0.9,\rowA) {$d_i^{(1)}$};
\draw[dotted, primary]
    (\tAa,{\rowA+0.3}) -- (\tAb,{\rowA+0.6}) -- (\tAc,{\rowA+0.4}) --
    (\tAf,{\rowA+0.8});
\foreach \t/\v in {\tAa/0.3, \tAb/0.6, \tAc/0.4, \tAf/0.8}{
    \fill[primary] (\t,{\rowA+\v}) circle (1.4pt);
    \draw[primary!50] (\t,-0.42) -- (\t,-0.38);
}

\node[left, font=\small] at (-0.9,\rowB) {$d_i^{(2)}$};
\draw[dotted, secondary]
    (\tBa,{\rowB-0.2}) -- (\tBb,{\rowB+0.3}) -- (\tBc,{\rowB+0.5}) --
    (\tBd,{\rowB-0.3}) -- (\tBe,{\rowB+0.2}) -- (\tBf,{\rowB+0.4});
\foreach \t/\v in {\tBa/-0.2, \tBb/0.3, \tBc/0.5, \tBd/-0.3, \tBe/0.2, \tBf/0.4}{
    \fill[secondary] (\t,{\rowB+\v}) circle (1.4pt);
    \draw[secondary!50] (\t,-0.42) -- (\t,-0.38);
}

\node[left, font=\small] at (-0.9,\rowC) {$d_i^{(3)}$};
\draw[dotted, accent]
    (\tCa,{\rowC+0.1}) -- (\tCb,{\rowC+0.1}) --
    (\tCd,{\rowC+0.9}) -- (\tCe,{\rowC+0.2}) -- (\tCf,{\rowC+0.1});
\foreach \t/\v in {\tCa/0.1, \tCb/0.1, \tCd/0.9, \tCe/0.2, \tCf/0.1}{
    \fill[accent] (\t,{\rowC+\v}) circle (1.4pt);
    \draw[accent!50] (\t,-0.42) -- (\t,-0.38);
}

\draw[decorate, decoration={brace, amplitude=6pt}, thick]
    (6.3,{\rowA+0.9}) -- (6.3,{\rowC-0.2})
    node[midway, right=8pt, font=\small, align=left]
    {$d_i = (d_i^{(1)}, d_i^{(2)}, d_i^{(3)})$\\[2pt]
     $c=3$\\[2pt]
     };

\end{tikzpicture}
\caption{An irregular multivariate time series $d_i \in
\mathcal{D}$ with $c=3$ channels, each observed on its own,
independent timestamp grid. Note that no timestamp is shared
across the three channels. Both the spacing and the number of
observations vary from one channel to another, and this
irregularity is compounded across series $d_i \in \mathcal{D}$.
Reconciling these independent grids onto a common timeline
(Section~\ref{subsec:ind-inst}) is a necessary preprocessing
step prior to signature computation.}
\label{fig:irregular_mts}
\end{figure}
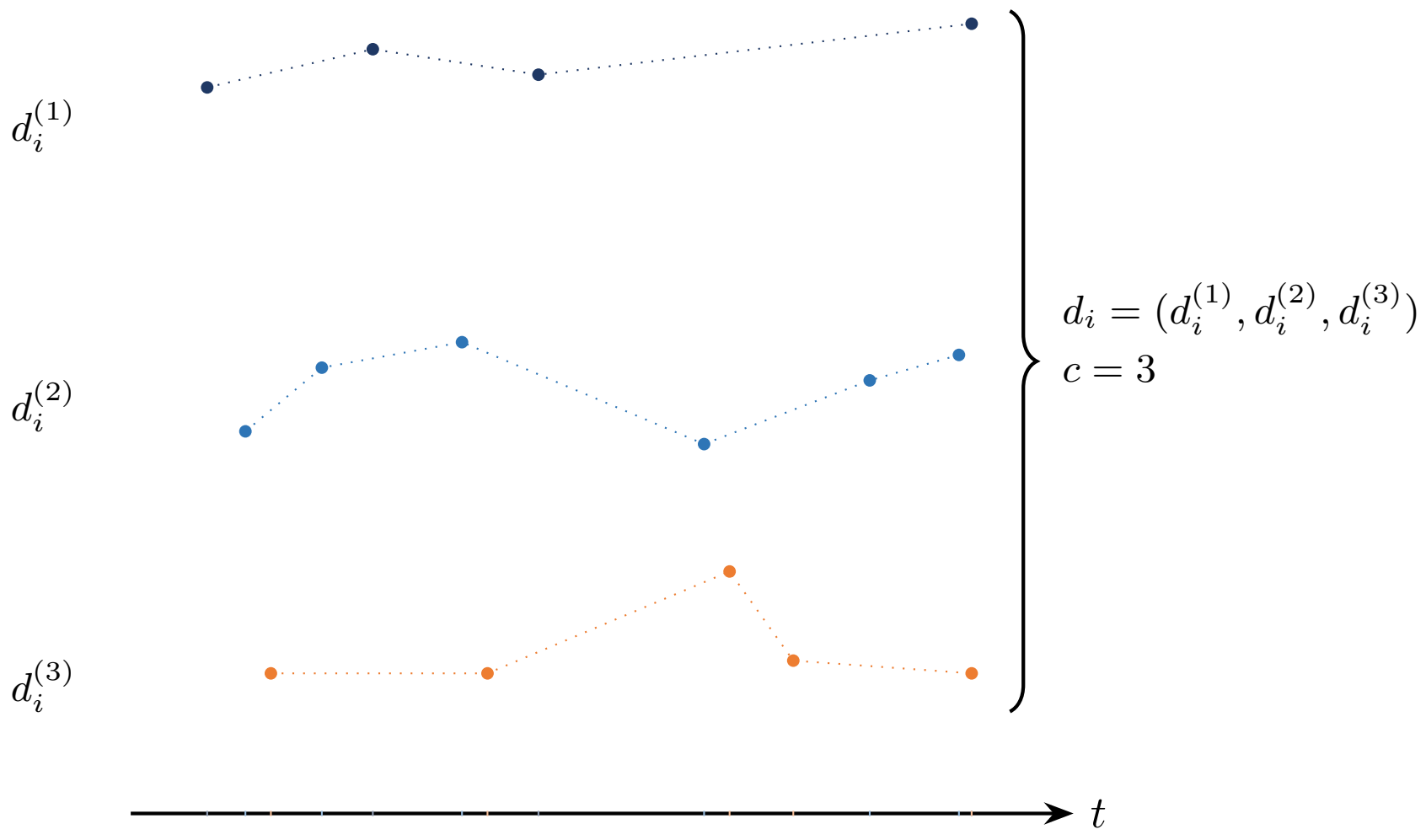

\paragraph{Setting.}[see Figure~\ref{fig:irregular_mts}]
Let $\mathcal{D} = \{d_1, \ldots, d_n\}$ be a dataset of $n$
irregular multivariate time series with $c$ channels. Each
series $d_i$ is a collection of $c$ univariate series, each
observed on its own timestamp grid:
\[
    d_i = \left(d_i^{(1)}, \ldots, d_i^{(c)}\right),
    \qquad
    d_i^{(j)} : \left\{t_{0,i}^{(j)}, \ldots, t_{f_j,i}^{(j)}\right\}
    \longrightarrow \mathbb{R},
    \quad j = 1, \ldots, c.
\]
No assumption is made on the timestamp grids
$\{t_{0,i}^{(j)}, \ldots, t_{f_j,i}^{(j)}\}$: they may be
irregularly spaced, differ in cardinality $f_j+1$ from one
channel to another, and are not required to coincide across
channels $j$: a series is not, in general, jointly observed on
a common grid. A supervised learning task is attached to
$\mathcal{D}$: each series carries a label among $Q$ classes, and a classifier is to
be trained on annotated series.

\paragraph{Problem.}
The labeling of such series needs, in general, a huge amount of human expert effort, particularly for large training datasets that can range up to millions of points to annotate. The main objective is to considerably reduce that amount of effort by making it possible to label fewer series without hindering the classification performance. The other objective is to increase this classification performance by making the model train only on the most relevant data, i.e.\ data that carries the specific and important information on which the classification task is mainly based. In summary, the problem we are trying to solve is: among a large number of unlabeled points, what proportion of this total mass carries the relevant information, and how can we select the points that, once labeled, would at least preserve, if not increase, the classification performance of a model trained on them?

\paragraph{Distributional point of view.}
A dataset is an empirical sample from an underlying data
distribution, and learning algorithms seek to minimize an expected
loss against this distribution, not against the sample itself. The
effectiveness of a training subset should therefore be measured
against the distribution rather than the sample: a good coreset is
one whose empirical distribution is close to that of the full
dataset~\cite{claici}. Formally, given a feature map
$\Phi : \mathcal{D} \to \mathbb{R}^{d}$ (constructed in
Section~\ref{subsec:signatures}), the dataset and a candidate
coreset $\mathcal{C} \subseteq \mathcal{D}$ of cardinality $k$
induce empirical measures on $\mathbb{R}^d$:
\[
    \hat{\mu}_{\mathcal{D}}
    = \frac{1}{n} \sum_{i=1}^{n} \delta_{\Phi(d_i)},
    \qquad
    \hat{\mu}_{\mathcal{C}}
    = \frac{1}{k} \sum_{d \in \mathcal{C}} \delta_{\Phi(d)}
\]
where $\delta_x$ denotes the Dirac measure at $x \in \mathbb{R}^d$.

\paragraph{Central hypothesis.}
Our working hypothesis is that the distributional discrepancy
between $\hat{\mu}_{\mathcal{C}}$ and $\hat{\mu}_{\mathcal{D}}$
in a well-chosen representation space is a reliable proxy for the
downstream performance gap between a model trained on $\mathcal{C}$
and one trained on $\mathcal{D}$:
\begin{equation}
    \label{eq:hypothesis}
    \text{minimizing } \mathsf{d}\!\left(\hat{\mu}_{\mathcal{C}},
    \hat{\mu}_{\mathcal{D}}\right)
    \;\approx\;
    \text{minimizing } \mathsf{d}\!\left(
    \mathrm{perf}_{\mathcal{C}}{(model)}, \mathrm{perf}_{\mathcal{D}}{(model)}\right).
\end{equation}
We do not claim a formal equivalence for general nonlinear
classifiers; rather, Eq.~\eqref{eq:hypothesis} is stated as a design
principle whose validity is assessed empirically in
Section~\ref{sec:experiments}, where we measure the correlation
between the Sliced Wasserstein distance of the selected coresets
and the degradation of downstream classification metrics.

\paragraph{Coreset selection problem.}
For a fixed annotation budget $k \ll n$, the optimal coreset is
defined as the coreset which minimizes the Sliced Wasserstein distance between the two empirical distributions
\begin{equation}
    \label{eq:coreset_problem}
    \mathcal{C}^{\star}
    = \underset{\substack{\mathcal{C} \subseteq \mathcal{D} \\
      |\mathcal{C}| = k}}{\arg\min}\;
      \mathrm{SW}_p\!\left(\hat{\mu}_{\mathcal{C}},\,
      \hat{\mu}_{\mathcal{D}}\right),
\end{equation}
where $\mathrm{SW}_p$ denotes the Sliced Wasserstein distance of
order $p$ (Section~\ref{subsec:sw}). Our hypothesis can be reformalized as:
\begin{equation}
    \label{eq:hypothesis2}
    \text{minimizing } \mathrm{SW}_p\!\left(\hat{\mu}_{\mathcal{C}},\,
      \hat{\mu}_{\mathcal{D}}\right)
    \;\approx\;
    \text{minimizing } \mathsf{d}\!\left(
    \mathrm{perf}_{\mathcal{C}}{(model)}, \mathrm{perf}_{\mathcal{D}}{(model)}\right).
\end{equation}

\subsection{Signature Representation}
\label{subsec:signatures}

Solving Eq.~\eqref{eq:coreset_problem} requires a distance between
distributions of time series. Optimal transport distances are
naturally defined over measures on a metric space~\cite{computationalot}; the difficulty is
that irregularly sampled series of variable lengths do not embed
canonically into a common metric space. 

Classical remedies each address only part of the problem posed
by irregular multivariate time series. Interpolation onto a
regular grid, and by extension methods built on top of it such
as wavelets~\citep{wavelets} or Gaussian
processes~\citep{gaussian}, require fixing an output
grid resolution as a hyperparameter, trading off information loss
against a dimensionality that grows with the sampling rate and
series duration. Elastic distances such as
DTW~\citep{dtw} and LCSS~\citep{lcss} avoid this trade-off by
aligning series directly on their observed timestamps, but the
resulting dissimilarities are not proper metrics and yield only a
pairwise score rather than a vector representation usable in a
distributional objective. Gaussian processes~\citep{gaussian} are also
costly to fit at the scale of a large dataset and do not yield a
canonical fixed-length vector without an additional, non-canonical
reduction step. Learnt embeddings~\citep{embedding} require a prior
training phase, are not interpretable, and offer no guarantee of
preserving the geometric structure of the underlying paths. 

To address these limitations, we instead rely on the \emph{path
signature} from rough path theory~\citep{lyons1998differential,
chevyrev2016primer}, which maps any path of bounded variation to a
fixed-dimensional vector while retaining, up to truncation, the full
information of the path.
This is especially relevant when what drives classification is not the isolated per-dimension values, but how the dimensions jointly evolve, their correlations and the shape traced out by their trajectories. 
The signature transform also relies on
an interpolation of the discrete observations into a continuous
path (Section~\ref{subsec:ind-inst}), and does not eliminate
this modeling choice altogether. However, by the reparametrization
invariance of Proposition~\ref{prop:reparam}, this interpolation
step is only a technical device: the resulting signature is
insensitive to the timing chosen \emph{within} each interval
between consecutive observations, and its dimension is fixed
independently of the sampling rate or series length. \\
We begin by the definition of the truncated signature.

\begin{definition}[Truncated signature]
Let $X : \{t_0, \ldots, t_f\} \to \mathbb{R}^{c}$ be a discrete
path, extended to continuous time by an interpolation scheme
(Section~\ref{subsec:ind-inst} discusses this choice). For a
multi-index $(j_1, \ldots, j_m) \in \{1, \ldots, c\}^m$, the
corresponding signature coefficient is the iterated integral
\[
    S(X)^{(j_1, \ldots, j_m)}
    = \idotsint\displaylimits_{t_0 < s_1 < \cdots < s_m < t_f}
      \mathrm{d}X^{j_1}_{s_1} \cdots \mathrm{d}X^{j_m}_{s_m},
\]
approximated on the discrete path by the Riemann sum
\[
    S(X)^{(j_1, \ldots, j_m)}
    = \sum_{s_1 = t_0}^{t_{f-1}} \Delta X^{j_1}_{s_1}
      \sum_{s_2 > s_1} \Delta X^{j_2}_{s_2}
      \cdots
      \sum_{s_m > s_{m-1}} \Delta X^{j_m}_{s_m},
    \qquad
    \Delta X^{j}_{s} = X^{j}_{s^{+}} - X^{j}_{s},
\]
where $s^{+}$ denotes the timestamp following $s$. The
\emph{truncated signature of order $N$} collects all coefficients up
to level $N$:
\[
    S^{N}(X)
    = \left(1,\;
      S(X)^{(1)}, \ldots, S(X)^{(d)},\;
      S(X)^{(1,1)}, \ldots, S(X)^{(d,d)},\;
      \ldots,\;
      S(X)^{(j_1, \ldots, j_N)}\right)
    \in \mathbb{R}^{m_{N,c}},
\]
with $m_{N,c} = \sum_{m=0}^{N} c^{m} = \frac{c^{N+1}-1}{c-1}$.
\end{definition}

\begin{example}[Signature of a 3-point path, up to order 2]
\label{ex:three_point}
Let $X : \{t_1, t_2, t_3\} \to \mathbb{R}^{c}$ be a discrete path
with $X(t_1) = x$, $X(t_2) = y$, $X(t_3) = z$, and let
\[
    \Delta_1 = y - x, \qquad \Delta_2 = z - y
\]
denote the two increments. The piecewise linear interpolation of
$X$ consists of two segments, on which $X$ moves at constant
speed $\Delta_1$ over $[t_1,t_2]$ and $\Delta_2$ over $[t_2,t_3]$.
We compute the signature directly from the defining iterated
integrals.

\paragraph{Level 0.}
\[
    S^{()} = 1.
\]

\paragraph{Level 1.} For $i \in \{1, \ldots, c\}$,
\[
    S^{(i)} = \int_{t_1}^{t_3} dX^i_t
    = \Delta_1^i + \Delta_2^i = z^i - x^i,
\]
depending only on the endpoints of the path.

\paragraph{Level 2.} For $i, j \in \{1, \ldots, c\}$,
\[
    S^{(i,j)} = \int_{t_1 < s < t < t_3} dX_s^i\, dX_t^j.
\]
Splitting the domain of integration according to which segment
each of $s, t$ belongs to gives three disjoint contributions.

\emph{Both $s,t$ in segment 1} (i.e.\ $t_1 < s < t < t_2$): on
this segment the path moves at constant velocity,
$dX_u^i = \frac{\Delta_1^i}{t_2-t_1}\, du$ for $u \in [t_1,t_2]$,
so
\[
    \int_{t_1 < s < t < t_2} dX_s^i\, dX_t^j
    = \frac{\Delta_1^i \Delta_1^j}{(t_2-t_1)^2}
      \int_{t_1 < s < t < t_2} ds\, dt
    = \frac{\Delta_1^i \Delta_1^j}{(t_2-t_1)^2}
      \cdot \frac{(t_2-t_1)^2}{2}
    = \tfrac{1}{2}\Delta_1^i \Delta_1^j,
\]
where the factor $(t_2-t_1)^2$ cancels exactly between the
squared velocity and the area of the ordering simplex.

\emph{Both $s,t$ in segment 2}: by the same computation on
$[t_2,t_3]$,
\[
    \int_{t_2 < s < t < t_3} dX_s^i\, dX_t^j
    = \tfrac{1}{2}\Delta_2^i \Delta_2^j.
\]

\emph{Finally $s$ in segment 1, $t$ in segment 2} (the only way to have
$s < t$ with $s,t$ in different segments, since segment 1
entirely precedes segment 2):
\[
    \int_{t_1}^{t_2} dX_s^i
    \int_{t_2}^{t_3} dX_t^j
    = \Delta_1^i \Delta_2^j.
\]

Summing the three contributions,
\begin{equation}
    \label{eq:level2_threepoint}
    S^{(i,j)}
    = \underbrace{\tfrac{1}{2}\Delta_1^i \Delta_1^j}_{\text{self, seg.\ 1}}
    \;+\;
    \underbrace{\Delta_1^i \Delta_2^j}_{\text{cross-term}}
    \;+\;
    \underbrace{\tfrac{1}{2}\Delta_2^i \Delta_2^j}_{\text{self, seg.\ 2}}.
\end{equation}
The self-interaction terms are symmetric in $(i,j)$, while the
cross-term is the only contribution carrying information about
the relative ordering of the two segments.

\paragraph{Lévy area.} Subtracting Eq.~\eqref{eq:level2_threepoint}
from its transpose, the symmetric self-interaction terms cancel
exactly, leaving
\[
    S^{(i,j)} - S^{(j,i)}
    = \Delta_1^i \Delta_2^j - \Delta_1^j \Delta_2^i,
\]
the signed area swept between the two consecutive increments.
\end{example}

\vspace{1em}
Three properties make the signature suitable for our purpose.
First, it is invariant under time reparametrization, hence
insensitive to the sampling rate of the underlying path, only the
trajectory matters.

\begin{proposition}[Invariance to time reparametrization~\cite{chevyrev2016primer}]
\label{prop:reparam}
Let $X : [0,1] \to \mathbb{R}^{c}$ be a continuous piecewise
smooth path. Let $\psi : [0,1] \to [0,1]$ be continuously
differentiable, increasing, and surjective. Then
\[
    S^{N}(X) = S^{N}(X \circ \psi), \qquad \forall N \geq 0.
\]
\end{proposition}

Second, by Lyons' uniqueness theorem, the signature characterizes a path up to tree-like equivalence.
\begin{proposition}[Uniqueness of signature~\cite{boedi, hambly}]
\label{prop:tree-like}
Let $X, Y : [0,T] \to \mathbb{R}^{c}$ be two paths of bounded
variation. Then
\[
    S(X) = S(Y) \quad \Longleftrightarrow \quad
    X \sim Y,
\]
where $\sim$ denotes \emph{tree-like equivalence}. We refer the reader to Hambly~\cite{hambly} for the precise definition of tree-like equivalence.
\end{proposition}

Recovering properties of a path from its signature remains a challenging
inverse problem in general; see Lyons-Xu~\cite{lyon-xu, lyon-xu2} and Geng~\cite{geng}. However, since our paths are time-augmented, this equivalence is trivial on our inputs: a strictly increasing time coordinate rules out any non-trivial retracement. 

\begin{definition}[Time augmentation]
\label{def:time_aug}
Let $X : \{t_0, \ldots, t_f\} \to \mathbb{R}^{c}$ be a discrete path. The \emph{time-augmented} path
$\tilde{X} : \{t_0, \ldots, t_f\} \to \mathbb{R}^{c+1}$ is defined by
prepending the (normalized) timestamp as an additional
coordinate:
\[
    \tilde{X}_{t_k} = (t_k,\, X_{t_k}), \qquad k = 0, \ldots, f.
\]
\end{definition}

\begin{remark}
Time augmentation removes the reparametrization invariance of
Proposition~\ref{prop:reparam}: since the augmented coordinate is
strictly increasing, the signature $S^N(\tilde{X})$ becomes sensitive
to the timing at which variations occur, not only to the shape of
the trajectory in $\mathbb{R}^c$. It also reduces the
tree-like equivalence of Proposition~\ref{prop:tree-like}
to a mere translation equivalence.
\end{remark}

\begin{proposition}[Uniqueness of signature, time-augmented case~\cite{hambly,chevyrev2016primer}]
\label{prop:unique_translation}
Let $X : [a,b] \to \mathbb{R}^{c}$ be a continuous path of bounded variation and let $\tilde{X}$ denote its time-augmented lift
(Definition~\ref{def:time_aug}). Then $S(\tilde{X})$ uniquely
determines $X$ up to translation.
\end{proposition}

The only information loss in our construction is therefore the truncation at order 
N, whose magnitude is controlled by the classical factorial decay bound. 

\begin{proposition}[Factorial decay~\cite{chevyrev2016primer}]
\label{prop:decay}
Let $X$ be a path of bounded variation with
$\|X\|_{1\text{-var}} \leq M$. Then for any multi-index
$(j_1,\ldots,j_m)$,
\[
    \left|S^{(j_1,\ldots,j_m)}(X)\right| \leq \frac{M^m}{m!}
\]
\end{proposition}

Third, a
universality property holds: any continuous functional of the path
can be approximated arbitrarily well by a \emph{linear} functional
of its truncated signature for $N$ large enough.

\begin{proposition}[Universal nonlinearity~\cite{bonnier}]
\label{prop:universality}
Let $F$ be a real-valued continuous function on continuous
piecewise smooth paths in $\mathbb{R}^{c}$, and let $\mathcal{K}$
be a compact set of such paths, with $X_0 = 0$ for all
$X \in \mathcal{K}$ (to remove translation invariance). Let
$\varepsilon > 0$. Then there exists a linear functional $L$
such that
\[
    \bigl|F(X) - L\bigl(S(X)\bigr)\bigr| < \varepsilon,
    \qquad \forall X \in \mathcal{K}.
\]
\end{proposition}

Level-1 terms
capture total displacements; level-2 terms capture pairwise signed
areas (Lévy areas), which encode the \emph{order} in which
components move relative to each other; and higher order terms capture progressively finer interactions among three or more components, information that, by Proposition~\ref{prop:decay}, carries vanishing magnitude as the order grows.

\paragraph{Rescaling.}
The level-$m$ coefficients of a signature decay factorially, see Proposition~\ref{prop:decay}. Left uncorrected, this
decay makes higher-order terms numerically negligible in any
Euclidean computation over the signature vector. We therefore apply
the component-wise rescaling map
\[
    \rho\!\left(S^{(j_1, \ldots, j_m)}\right)
    = m! \cdot S^{(j_1, \ldots, j_m)},
\]
which brings all levels to comparable magnitude
$O(M^{m})$ making the Euclidean distance on
signature vectors meaningful.

\begin{lemma}[Boundedness of rescaled signatures]
\label{lem:bounded}
Let $X$ be a path with total variation bounded by $M$. Then every
rescaled signature coefficient satisfies
$|\rho(S^{(j_1,\ldots,j_m)}(X))| \leq M^{m}$, and consequently
$\|\rho(S^{N}(X))\|_2 \leq R(N, c, M)$ with
$R(N,c,M) = \big(\sum_{m=0}^{N} c^{m} M^{2m}\big)^{1/2}$.
\end{lemma}

\begin{remark}[Truncation versus rescaling]
The factorial decay of Proposition~\ref{prop:decay} justifies
truncation at order $N$: it shows that the magnitude of the
omitted tail ($m > N$) vanishes rapidly, so that most of the
signature's total mass is retained by a moderate $N$. This
same decay, however, also occurs \emph{within} the retained
levels $0, \ldots, N$; level $N$ is systematically smaller
than level $1$ by a combinatorial artifact of the iterated
integral, not
because it carries less discriminative information. Left
uncorrected, this would cause any Euclidean-based computation
on the truncated vector to be dominated by low-order terms.
Rescaling corrects this
imbalance \emph{among} the retained levels, and is therefore
natural with the truncation
decision.
\end{remark}

\paragraph{Generic representation.}
In its general form, the SOTS representation of a series $d_i$ is
\[
    \Phi(d_i) = \rho\!\left(S^{N}(\tilde{X}_i)\right)
    \in \mathbb{R}^{d},
    \qquad d = m_{N, c+1},
\]
possibly concatenated with a small number of scalar descriptors
(the industrial instantiation of Section~\ref{subsec:ind-inst}
adds one such descriptor and splits the channels into two separate
paths). The truncation order $N$ is a hyperparameter trading
expressiveness against dimensionality; we use $N = 2$ throughout
and discuss this choice in Section~\ref{subsec:ind-inst}.

\subsection{Sliced Wasserstein Distance}
\label{subsec:sw}

In this section, we motivate the use of the Sliced Wasserstein (SW) distance between the empirical distributions. We first begin by the definition of the Wasserstein distance for discrete measures.

\begin{definition}[Wasserstein distance~\cite{computationalot}]
Let $p \geq 1$ and let $a, b$ be two discrete probability measures
supported on $\{x_1, \ldots, x_n\}$ and $\{y_1, \ldots, y_m\}$
respectively. The $p$-Wasserstein distance is
\[
    W_p(a, b)
    = \left(
      \min_{P \in U(a,b)}
      \sum_{i,j} \|x_i - y_j\|^{p}\, P_{ij}
      \right)^{1/p}
    \qquad
    U(a,b) = \left\{ P \in \mathbb{R}_{+}^{n \times m}
    \,\middle|\, P \mathds{1}_m = a,\; P^{\top} \mathds{1}_n = b
    \right\},
\]
where $P_{ij}$ represents the mass transported from $x_i$ to $y_j$.
\end{definition}

\begin{definition}[Wasserstein distance, general case~\cite{computationalot}]
Let $p \geq 1$ and let $\mu, \nu$ be two probability measures on
a Polish space $(\mathcal{X}, d)$ (i.e., a separable, completely metrizable topological space) with finite $p$-th moment. The
$p$-Wasserstein distance is
\[
    W_p(\mu, \nu)
    = \left(
      \inf_{\pi \in \Pi(\mu,\nu)}
      \int_{\mathcal{X} \times \mathcal{X}}
      d(x, y)^{p} \, \mathrm{d}\pi(x, y)
      \right)^{1/p},
\]
where
\[
    \Pi(\mu, \nu)
    = \left\{ \pi \in \mathcal{P}(\mathcal{X} \times \mathcal{X})
    \,\middle|\, (\mathrm{pr}_1)_{\#}\pi = \mu,\;
    (\mathrm{pr}_2)_{\#}\pi = \nu \right\}
\]
is the set of couplings with marginals $\mu$ and $\nu$, and
$\mathrm{pr}_1, \mathrm{pr}_2$ denote the projections onto the
first and second coordinates. See Villani~\cite{villani} for a detailed study of $W_p$.
\end{definition}

\begin{remark}
The discrete formulation given previously is the special case
where $\mu = \sum_i a_i \delta_{x_i}$ and
$\nu = \sum_j b_j \delta_{y_j}$ are finitely supported: the
coupling $\pi$ reduces to the transport plan matrix $P$, with
$\pi = \sum_{i,j} P_{ij}\, \delta_{(x_i, y_j)}$, and the integral
becomes the finite sum $\sum_{i,j} \|x_i - y_j\|^p P_{ij}$.
\end{remark}

Computing $W_p$ exactly requires
solving a linear program at cost $O(n^{3} \log n)$, which is
prohibitive when the distance is evaluated repeatedly inside a
selection loop. Another relevant result is the closed form solution of this minimization problem in one dimension.

\begin{proposition}[Closed form in one dimension~{\cite[Remark~2.30]{computationalot}}]
$W_p$ admits
a closed form in dimension one. For two measures $\alpha, \beta$ on
$\mathbb{R}$ with quantile functions
$F_\alpha^{-1}, F_\beta^{-1}$,
\begin{equation}
    \label{eq:w1d}
    W_p^{p}(\alpha, \beta)
    = \int_{0}^{1}
      \left| F_\alpha^{-1}(t) - F_\beta^{-1}(t) \right|^{p}
      \mathrm{d}t,
\end{equation}
\end{proposition}
computable in $O(n \log n)$ by sorting~{\cite[Sec.~1.1]{peyre2019coursenotes}}.\\

The Sliced Wasserstein distance exploits this fact by projecting multi-dimensional measures onto random directions to benefit from the fast computation for uni-dimensional measures.

\begin{definition}[Sliced Wasserstein~\cite{sw}]
For measures on $\mathbb{R}^{d}$, one projects onto directions
$\theta \in \mathbb{S}^{d-1}$ in the unit sphere and averages the resulting
one-dimensional distances:
\[
    \mathrm{SW}_p\!\left(\hat{\mu}_{\mathcal{C}},
    \hat{\mu}_{\mathcal{D}}\right)
    = \left(
      \int_{\mathbb{S}^{d-1}}
      W_p^{p}\!\left(\theta_{\#}\hat{\mu}_{\mathcal{C}},\,
      \theta_{\#}\hat{\mu}_{\mathcal{D}}\right)
      \mathrm{d}\sigma(\theta)
      \right)^{1/p},
\]
where $\sigma$ is the uniform measure on $\mathbb{S}^{d-1}$ and
$\theta_{\#}\hat{\mu} = \frac{1}{|\mathcal{C}|}
\sum_{d \in \mathcal{C}} \delta_{\langle \theta, \Phi(d) \rangle}$
is the push-forward of $\hat{\mu}$ along $\theta$.
\end{definition}

The
$\mathrm{SW}_p$ is a proper metric on probability measures with
finite $p$-th moment, and vanishes if and only if the two measures
coincide. For general arbitray measures, the following result is established~\cite{sw}:

\begin{proposition}[Sliced Wasserstein is dominated by Wasserstein]
\label{prop:sw_leq_w}
Let $\mu, \nu$ be two probability measures on $\mathbb{R}^{d}$ with
finite $p$-th moment, $p \geq 1$. Then
\[
    \mathrm{SW}_p(\mu, \nu) \;\leq\; W_p(\mu, \nu).
\]
This is a classical consequence of the fact that any linear
projection is 1-Lipschitz; see, e.g.,~\cite[Theorem~1]{w-sw} for the sharper
form with an explicit constant $C(d,p) \leq 1$, or~\cite{sw} for the unconstrained version used
throughout the Sliced Wasserstein literature. We include a
self-contained proof below for completeness.
\end{proposition}

\begin{proof}
Fix a direction $\theta \in \mathbb{S}^{d-1}$ and let
$P_\theta : \mathbb{R}^{d} \to \mathbb{R}$, $x \mapsto \langle
\theta, x \rangle$ denote the corresponding projection. Since
$\|\theta\|_2 = 1$, the Cauchy–Schwarz inequality gives, for any
$x, y \in \mathbb{R}^{d}$,
\[
    |P_\theta(x) - P_\theta(y)|
    = |\langle \theta, x - y \rangle|
    \leq \|\theta\|_2 \, \|x - y\|_2
    = \|x - y\|_2,
\]
so $P_\theta$ is $1$-Lipschitz.

Let $\gamma^\star \in \Gamma(\mu, \nu)$ be an optimal transport
plan attaining $W_p(\mu,\nu)$, i.e.
\[
    W_p^{p}(\mu, \nu)
    = \int_{\mathbb{R}^{d} \times \mathbb{R}^{d}}
      \|x - y\|_2^{p} \, \mathrm{d}\gamma^\star(x,y).
\]
The pushforward $(P_\theta, P_\theta)_{\#} \gamma^\star$ of
$\gamma^\star$ under the map $(x,y) \mapsto (P_\theta(x),
P_\theta(y))$ is a coupling between the marginals
$\theta_{\#}\mu$ and $\theta_{\#}\nu$, and therefore belongs to
the admissible set $\Gamma(\theta_{\#}\mu, \theta_{\#}\nu)$ for
the one-dimensional transport problem. By optimality of $W_p$
over all admissible couplings,
\[
    W_p^{p}(\theta_{\#}\mu, \theta_{\#}\nu)
    \;\leq\;
    \int_{\mathbb{R} \times \mathbb{R}}
    |u - v|^{p} \,
    \mathrm{d}\bigl[(P_\theta, P_\theta)_{\#}\gamma^\star\bigr](u,v)
    =
    \int_{\mathbb{R}^{d} \times \mathbb{R}^{d}}
    |P_\theta(x) - P_\theta(y)|^{p} \,
    \mathrm{d}\gamma^\star(x,y).
\]
Using the Lipschitz bound above,
\[
    W_p^{p}(\theta_{\#}\mu, \theta_{\#}\nu)
    \;\leq\;
    \int_{\mathbb{R}^{d} \times \mathbb{R}^{d}}
    \|x - y\|_2^{p} \, \mathrm{d}\gamma^\star(x,y)
    = W_p^{p}(\mu, \nu).
\]
This bound holds for \emph{every} $\theta \in \mathbb{S}^{d-1}$,
and in particular does not depend on $\theta$ on the right-hand
side. Integrating both sides over $\theta \sim \sigma$, the
uniform measure on $\mathbb{S}^{d-1}$,
\[
    \mathrm{SW}_p^{p}(\mu, \nu)
    = \int_{\mathbb{S}^{d-1}}
      W_p^{p}(\theta_{\#}\mu, \theta_{\#}\nu) \,
      \mathrm{d}\sigma(\theta)
    \;\leq\;
    \int_{\mathbb{S}^{d-1}}
      W_p^{p}(\mu, \nu) \, \mathrm{d}\sigma(\theta)
    = W_p^{p}(\mu, \nu).
\]
Taking the $p$-th root on both sides yields the claim.
\end{proof}

\paragraph{Exact computation with unequal sample sizes.}
In our setting the two projected measures have different supports of
sizes $k = |\mathcal{C}|$ and $n = |\mathcal{D}|$. Let
$u_1, \ldots, u_k$ and $v_1, \ldots, v_n$ denote the projected
samples for a fixed $\theta$. Both quantile functions
$F_{\mathcal{C}}^{-1}$ and $F_{\mathcal{D}}^{-1}$ are step
functions taking finitely many values, so the integral in
Eq.~\eqref{eq:w1d} can be computed \emph{exactly}. Let
$\tau_1 \leq \cdots \leq \tau_{k+n}$ be the sorted union of the rank
positions $\{i/k\}_{i=1}^{k} \cup \{j/n\}_{j=1}^{n}$ in the
quantile domain $(0,1]$; on each interval
$[\tau_\ell, \tau_{\ell+1})$ both quantile functions are constant,
and therefore
\begin{equation}
    \label{eq:breakpoints}
    W_p^{p}\!\left(\theta_{\#}\hat{\mu}_{\mathcal{C}},\,
    \theta_{\#}\hat{\mu}_{\mathcal{D}}\right)
    = \sum_{\ell=1}^{k+n-1}
      \left| F_{\mathcal{C}}^{-1}(\tau_\ell)
      - F_{\mathcal{D}}^{-1}(\tau_\ell) \right|^{p}
      (\tau_{\ell+1} - \tau_\ell),
\end{equation}
at total cost for sorting the rank positions $O\!\left((k+n)\log(k+n)\right)$ per direction,
with no discretization error.

\begin{remark}[Grid-based approximation]
Alternatively, the integral may be approximated on a grid of $M$
uniformly spaced points in $(0,1)$, typically $M = \max(k, n)$,
yielding a simpler implementation at the price of a discretization
error of order $O(1/M)$. We use the exact breakpoint formula
Eq.~\eqref{eq:breakpoints} throughout.
\end{remark}

\paragraph{Monte Carlo estimator and projection strategies.}
The spherical integral is estimated by Monte Carlo over $L$
directions:
\[
    \mathrm{SW}_p^{p}\!\left(\hat{\mu}_{\mathcal{C}},
    \hat{\mu}_{\mathcal{D}}\right)
    \approx \frac{1}{L} \sum_{\ell=1}^{L}
    W_p^{p}\!\left(\theta_{\ell\#}\hat{\mu}_{\mathcal{C}},\,
    \theta_{\ell\#}\hat{\mu}_{\mathcal{D}}\right),
    \qquad
    \theta_\ell \sim \mathrm{Unif}(\mathbb{S}^{D-1}),
\]
at total cost $O\!\left(L(k+n)\log(k+n)\right)$.\\

Signature vectors
are typically sparse (in our industrial data, approximately
half components are non-zero on average) so uniform directions may
waste projections on uninformative subspaces. We therefore also
implement three adaptive slicing strategies: the Energy-Based
Sliced Wasserstein (EBSW)~\citep{ebsw}, which reweights
uniform directions by their discriminative power, and the
Random-Path variants RPSW and IWRPSW~\citep{rpsw}, which draw directions
concentrated around normalized differences of samples from the two
measures via a Power Spherical distribution with concentration
parameter $\kappa$. As $\kappa \to 0$, RPSW recovers the standard
$\mathrm{SW}_p$, and IWRPSW converges to EBSW as $L \to \infty$.
The impact of the slicing strategy is evaluated in
Section~\ref{sec:experiments}.

\subsection{Coreset Selection Algorithms}
\label{subsec:algorithms}

Solving the coreset selection problem exactly, as defined in
Eq.~\eqref{eq:coreset_problem}, requires evaluating
$\binom{n}{k}$ candidate subsets and is therefore computationally
intractable for any realistic $n$; this motivates the greedy
heuristic of Algorithm~\ref{alg:greedy}. Independently of this
heuristic, however, one can ask a weaker but unconditional
question: does \emph{any} reasonable coreset of size $k$ become
an increasingly faithful distributional approximation of the
full dataset as $k$ grows? We answer this in the affirmative for
a uniformly random coreset, by adapting the empirical convergence
rate established by Fournier \& Guillin \cite{fournier} for the Wasserstein
distance to the signature representation space and to the Sliced
Wasserstein distance. The resulting guarantee decays at the
explicit rate $k^{-1/d}$, where $d$ is the dimension of the
signature vector (Section~\ref{subsec:signatures}). We state the
general theorem first, then derive its instantiation to our
setting.

\begin{theorem}[Empirical Wasserstein convergence rate,
{\cite[Theorem~1]{fournier}}]
\label{thm:fournier_guillin}
Let $d \geq 1$ and let $\mu \in \mathcal{P}(\mathbb{R}^{d})$,
where $\mathcal{P}(\mathbb{R}^{d})$ denotes the set of all
probability measures on $\mathbb{R}^{d}$. Let $0 < p < d/2$, and
assume that the $q$-th moment
\[
    M_q(\mu) := \int_{\mathbb{R}^{d}} \|x\|_2^{q} \, \mathrm{d}\mu(x)
\]
is finite for some $q > p$ with $q \neq d/(d-p)$. Let
$X_1, \ldots, X_N$ be i.i.d.\ random variables with law $\mu$,
and let $\mu_N = \frac{1}{N}\sum_{i=1}^{N} \delta_{X_i}$ be the
associated empirical measure. Define the (unnormalized) optimal
transport cost
\[
    T_p(\alpha, \beta)
    := \inf_{\pi \in \Pi(\alpha,\beta)}
    \int_{\mathbb{R}^{d} \times \mathbb{R}^{d}}
    \|x - y\|_2^{p} \, \mathrm{d}\pi(x,y),
\]
where $\Pi(\alpha,\beta)$ is the set of couplings of $\alpha$
and $\beta$, so that $T_p = W_p^{\,p}$ for $p > 1$ and
$T_p = W_p$ for $p \in (0,1]$. Then there exists a constant
$C(d,p,q) < \infty$, depending only on $d$, $p$, and $q$, such
that
\[
    \mathbb{E}\!\left[T_p(\mu_N, \mu)\right]
    \;\leq\; C(d,p,q)\, M_q(\mu)^{p/q}
    \left(N^{-p/d} + N^{-(q-p)/q}\right),
    \qquad \forall N \geq 1.
\]
\end{theorem}

The assumptions of Theorem~\ref{thm:fournier_guillin} are mild
and hold essentially automatically in our setting. The condition
$d > 2p$ is satisfied for any practical configuration of our
framework: for the common choice $p=2$ (squared Euclidean cost),
it reduces to $d > 4$, and the signature dimension $d$ grows
combinatorially with the truncation order $N$ and the
(time-augmented) path dimension; even the smallest non-trivial
configurations comfortably exceed this threshold, and our
industrial instantiation (Section~\ref{subsec:ind-inst}) gives
$d = 27$. The moment condition $M_q(\mu) < \infty$ is likewise
never a restriction in practice: any empirical distribution
supported on finitely many points has finite moments of every
order. This setting is therefore well suited to our context, and
we now state the resulting bound for the Sliced Wasserstein
distance between empirical measures of signature vectors.

\begin{proposition}[Distributional convergence of uniform sampling]
\label{prop:convergence}
Let $\hat\mu_{\mathcal{D}}$ denote the empirical measure of the
signature representation of the full dataset $\mathcal{D}$, and
let $(X_i)_{i \geq 1}$ be an i.i.d.\ sequence of
$\hat\mu_{\mathcal{D}}$-distributed random variables. The
empirical measure of a random coreset of size $k$ is
$\hat\mu_{\mathcal{C}} = \frac{1}{k}\sum_{i=1}^{k}
\delta_{X_i}$.

By Lemma~\ref{lem:bounded}, $\Phi(\mathcal{D})$ is supported in a
ball of radius $R = R(N,c,M)$ in $\mathbb{R}^{d}$, where
$d = m_{N,c+1}$, so $M_q(\hat\mu_{\mathcal{D}}) \leq R^{q}$ for
every $q > 0$. Assume $d > 2p$ and $p > 1$, and set
\[
    q = \frac{pd}{d-p} \;>\; p, \qquad
    q \neq \frac{d}{d-p} \;\;\text{(since $p \neq 1$)}.
\]
With this choice, $\dfrac{q-p}{q} = \dfrac{p}{d}$, so the two
terms of Theorem~\ref{thm:fournier_guillin} coincide. Recalling
$T_p = W_p^{\,p}$ for $p > 1$, Theorem~\ref{thm:fournier_guillin}
gives
\begin{equation}
    \label{eq:fg_applied}
    \mathbb{E}\!\left[
        W_p\!\left(\hat\mu_{\mathcal{C}}, \hat\mu_{\mathcal{D}}\right)^{p}
    \right]
    \;\leq\; 2\, C(d,p,q)\, R^{p}\, k^{-p/d}.
\end{equation}
By Lyapunov's inequality ($p > 1$) and since $\mathrm{SW}_p \leq W_p$
pointwise (Proposition~\ref{prop:sw_leq_w}),
\begin{equation}
    \label{eq:final_bound}
    \mathbb{E}\!\left[
        \mathrm{SW}_p\!\left(\hat\mu_{\mathcal{C}}, \hat\mu_{\mathcal{D}}\right)
    \right]
    \;\leq\;
    \mathbb{E}\!\left[W_p\!\left(\hat\mu_{\mathcal{C}},
    \hat\mu_{\mathcal{D}}\right)\right]
    \;\leq\;
    \left(\mathbb{E}\!\left[W_p\!\left(\hat\mu_{\mathcal{C}},
    \hat\mu_{\mathcal{D}}\right)^{p}\right]\right)^{1/p}
    \;\leq\; C'(d,p)\, R\, k^{-1/d},
\end{equation}
where $C'(d,p) = \left(2\, C(d,p,q)\right)^{1/p}$ depends only
on $d$ and $p$.
\end{proposition}

\begin{remark}
Theorem~\ref{thm:fournier_guillin} is stated for a general
measure $\mu$ on $\mathbb{R}^{d}$ and says nothing, on its own,
about signatures or coreset selection. Two properties specific to
our setting are what make it applicable here. First, the
boundedness established in Lemma~\ref{lem:bounded} guarantees
that the moment condition of Theorem~\ref{thm:fournier_guillin}
holds for \emph{every} order $q$ simultaneously, an unusually
favorable situation that lets us choose $q = \frac{pd}{d-p}$
freely so as to collapse the two terms of the general bound into
a single, explicit rate, rather than merely bounding one term
while leaving the other as a free parameter. Second, the
domination $\mathrm{SW}_p \leq W_p$ (Proposition~\ref{prop:sw_leq_w})
provides the bridge from this classical guarantee on the
Wasserstein distance to the Sliced Wasserstein objective that
SOTS actually optimizes. Taken together, these two observations
turn a general-purpose convergence theorem into a distributional
guarantee stated directly in terms of the signature truncation
order $N$: since $d = m_{N,c+1}$ grows combinatorially with $N$,
the rate $k^{-1/d}$ makes explicit, rather than merely
qualitative, the cost in statistical efficiency incurred by
increasing $N$, complementing the classification-accuracy
argument for keeping $N$ small discussed in
Section~\ref{subsec:uea}.
\end{remark}

Beyond random sampling, several established unsupervised coreset
selection criteria could, in principle, be applied in the
signature representation space. Herding \citep{herding, herding_2}
greedily selects the point that brings the coreset mean closest
to the dataset mean, $\arg\min_{d} \|\mu_{\mathcal{D}} -
\mathrm{mean}(\mathcal{C} \cup \{d\})\|_2$ which is a first-moment
matching criterion. Facility Location
\citep{fcltyloc, fcltyloc_2}
greedily maximizes a submodular coverage function built from
pairwise similarities, $\sum_i \max_{d \in \mathcal{C}}
s(\Phi(d_i), \Phi(d))$. $K$-means and $k$-medoids partition the
representation space into clusters and retain one representative
per cluster, minimizing within-cluster variance or pairwise
dissimilarity. None of these criteria is derived from a distance
between the coreset and dataset \emph{distributions}: they
either reduce the comparison to a single summary statistic (the
mean, for Herding), to pairwise point-to-point similarities
(Facility Location, $k$-medoids), or to a partition objective
with no direct relation to any measure-theoretic notion of
proximity ($k$-means). In particular, none of them involves an
optimal transport formulation, and their objectives are
insensitive to the higher-order geometric information that the
signature representation was specifically constructed to
capture, the signed areas and causal orderings encoded at level
two and beyond (Example~\ref{ex:three_point}). Applying these
criteria on top of signature features, as we do for comparison
in Section~\ref{sec:experiments}, therefore only partially
exploits the representation: the geometric structure is present
in $\Phi(\mathcal{D})$, but the selection criterion itself does
not use it beyond a pointwise or first-moment summary. We now
introduce SOTS-Greedy, which is, to our knowledge, the first
coreset selection procedure to combine the signature
representation with an explicit distributional objective: the
Sliced Wasserstein distance between empirical measures. The geometric information encoded at every retained signature
level is directly reflected in the selection criterion, rather
than discarded before it is used.

\paragraph{Greedy SW minimization (SOTS-Greedy).}
Starting from $\mathcal{C}_0 = \emptyset$, at each step $t$ the
point with the largest marginal reduction of the objective is
added:
\[
    d_t^{\star}
    = \underset{d \in \mathcal{D} \setminus \mathcal{C}_{t-1}}
      {\arg\min}\;
      \mathrm{SW}_p\!\left(
      \hat{\mu}_{\mathcal{C}_{t-1} \cup \{d\}},\,
      \hat{\mu}_{\mathcal{D}}\right),
    \qquad
    \mathcal{C}_t = \mathcal{C}_{t-1} \cup \{d_t^{\star}\},
\]
until $|\mathcal{C}_k| = k$. All remaining candidates at a given
step are evaluated in parallel on GPU, in chunks of size $B$ to bound memory
usage; the total complexity is
$O\!\left(k \cdot \lceil n/B \rceil \cdot L (k+n)\log(k+n)\right)$.
A by-product of the sequential construction is that the selection
order itself defines a family of nested coresets
$\mathcal{C}_1 \subset \cdots \subset \mathcal{C}_n$, so that
coresets of all intermediate sizes are obtained from a single run.\\

\begin{algorithm}[H]
\caption{SOTS-Greedy}
\label{alg:greedy}
\begin{algorithmic}[1]
\Require Dataset $\mathcal{X} = \{x_1, \ldots, x_n\}$, budget $k$,
number of projections $L$
\Ensure Coreset $\mathcal{C}$
\State $\mathcal{C} \leftarrow \emptyset$
\State $R \leftarrow \{1, \ldots, n\}$
\Comment{remaining candidates}
\For{$\mathrm{step} = 1, \ldots, k$}
    \For{each chunk $B \subseteq R$}
    \Comment{evaluated in parallel on GPU}
        \For{each $x \in B$}
            \State $\mathcal{C}_x \leftarrow \mathcal{C} \cup \{x\}$
            \State $s_x \leftarrow {SW}_p\!\left(
                \hat\mu_{\mathcal{C}_x}, \hat\mu_{\mathcal{X}}\right)$
        \EndFor
    \EndFor
    \State $x^\star \leftarrow \displaystyle
        \arg\min_{x \in R} s_x$
    \State $\mathcal{C} \leftarrow \mathcal{C} \cup \{x^\star\}$
    \State $R \leftarrow R \setminus \{x^\star\}$
\EndFor
\State \Return $\mathcal{C}$
\end{algorithmic}
\end{algorithm}

\begin{remark}[On comparing greedy to random sampling] Establishing
$\mathbb{E}[\mathrm{SW}_p(\mathcal{C}_k^{greedy})]\leq\mathbb{E}[\mathrm{SW}_p(\mathcal{C}_k^{\mathrm{unif}})]$ for
every $k$ by induction fails at the inductive step: while the
base case $k=1$ holds trivially (the greedy singleton minimizes
$\mathrm{SW}_p$ over \emph{all} singletons), the step from
$k-1$ to $k$ would require that a low value of
$\mathrm{SW}_p(\hat\mu_{\mathcal{C}_{k-1}}, \hat\mu_{\mathcal{D}})$
implies a favorable position for further improvement, a
monotonicity property that does not follow from the value alone
without additional structure such as submodularity, which we do
not establish here. We therefore do not claim an unconditional
theoretical advantage of SOTS-Greedy over random sampling, and
instead demonstrate it empirically in
Section~\ref{subsubsec:results1}.
\end{remark}

\begin{remark}
Sequential greedy constructions are known to achieve
rate-optimal asymptotic performance in the related setting of
$L^p$-optimal vector quantization against a continuous target
distribution, with freely positioned points and
Voronoi-optimal weights~\citep{quantization}. Whether an
analogous guarantee holds for SOTS-Greedy, constrained to
select points from a finite dataset under uniform weights, and
targeting a discrete rather than continuous reference measure,
is a structurally distinct question that we leave open for
future work.
\end{remark}

\subsection{Industrial Instantiation}
\label{subsec:ind-inst}

We now instantiate the framework on the industrial problem that
motivated this work: identifying false-positive discrepancies in
the regression testing of critical Instrumentation and Control
(I\&C) software for nuclear power plants.

\paragraph{Industrial context.}
Critical I\&C software executes highly safety-relevant functions (the emergency shutdown of a nuclear reactor, emergency braking
in trains, or the alerting of authorities to a potential threat)
encompassing every device from the initial sensing element to the
final control device. Such software must satisfy the safety
requirements imposed by international standards~\citep{standards}, which in turn demand a rigorous testing
process. In our setting, part of this process consists of
executing identical test scenarios in two environments: an
\emph{Oracle}, a theoretical environment encoding the behavior
expected from the specification of the tested module, and a
\emph{simulated} environment designed to reproduce real-world
operating conditions. Each output of a test scenario is a pair of
signals: a value and a binary flag indicating its validity, a
recurring notion across I\&C devices. The outputs of the two
environments must be compared, since any deviation from the
Oracle may reveal a defect in the software under test.

\paragraph{From discrepancies to false positives.}
Deviations between the Oracle and simulated outputs are flagged
by a Discrepancy Detection Module (DDM) and subsequently reviewed
by engineers. Not every flagged discrepancy, however, reflects an
actual software defect: limitations of the simulation
environment itself routinely produce discrepancies that do not
correspond to a fault (\emph{false positives}) for a variety of
distinct underlying causes, recurring across the report at
different timestamps. Distinguishing true defects from false
positives currently requires substantial manual review, and is
precisely the classification task this work aims to support with
a reduced, representative training set. Since a comparison
involves two outputs, each composed of a value and a validity
signal that may be sampled at different rates
(Figure~\ref{fig:irregular_mts}), this classification task is
naturally formulated as an Irregular Multivariate Time Series
Classification (IMTSC) problem, in which
capturing both temporal dependencies within a signal and
inter-variable dependencies across signals is essential. For more details on this industrial dataset and the classification method used, see~\cite{anas}.

\paragraph{Data structure.}
Each discrepancy $d_i$ is extracted around a detected gap
between two compared signals, and consists of four channels,
the value and the validity of each signal, organized into two
timestamp grids, one per signal, in the same spirit as the
general formulation of Section~\ref{subsec:problem}:
\[
    d_i = \bigl(d_i^{1},\, d_i^{2}\bigr),
    \qquad
    d_i^{j} : \{t_{0,i}^{j}, \ldots, t_{f_j,i}^{j}\}
    \longrightarrow \mathbb{R} \times \{0,1\},
    \quad
    t \longmapsto (x^{j}, v^{j}),
    \quad j=1,2,
\]
where $x^{j}$ is the (analog or binary) value of signal $j$ and
$v^{j} \in \{0,1\}$ its validity flag, jointly observed at each
timestamp of that signal's own grid. The two grids
$\{t_{0,i}^{1}, \ldots\}$ and $\{t_{0,i}^{2}, \ldots\}$ are not
assumed to coincide: the two compared signals are sampled
independently of one another, and in practice the second signal
is sampled at a higher rate than the first. The nature of a fault
is sometimes conditioned by the validity pattern rather than the
values alone, thus the two kinds of information carry a different
nature and should not be blended indiscriminately.

\begin{figure}[h]
    \centering
    \includegraphics[width=1\linewidth]{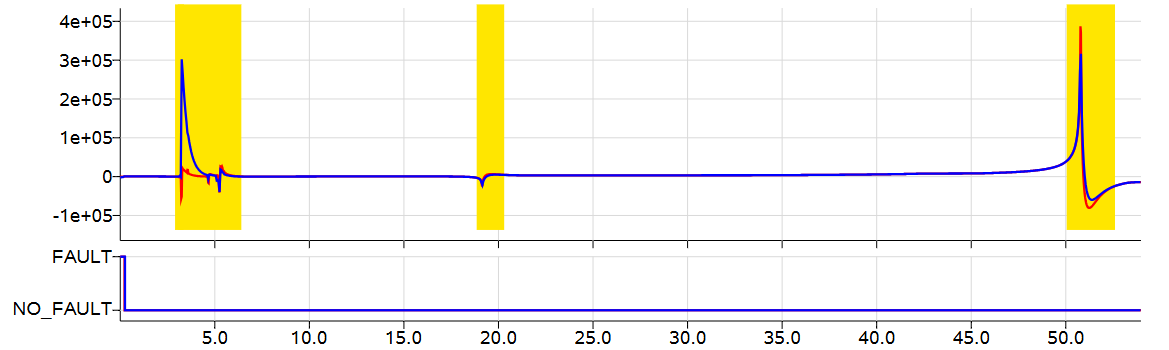}
    \caption{Example of a comparison between two outputs: the red line represents the
oracle output, while the blue one represents the simulation output. The top figure
shows the values, while the bottom figure shows the validity flags for each output. This
boolean flag is attached to each signal to indicate whether the value is significant or
not (for instance a degraded sensor that would still transmit a measurement, however
invalidated). The three yellow rectangles framing the three discrepancies are flagged by the DDM.}
    \label{fig:analog}
\end{figure}

\begin{figure}[h]
    \centering
    \includegraphics[width=1\linewidth]{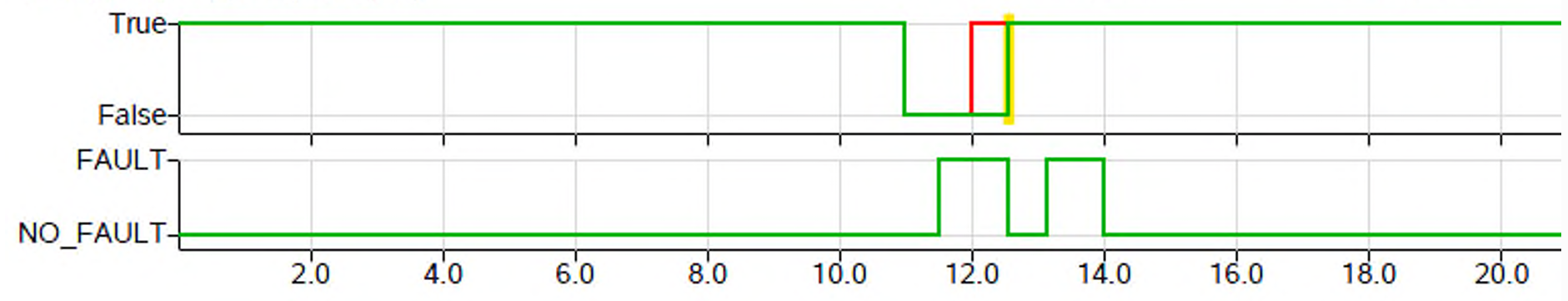}
    \caption{Example of a comparison between two outputs when the value of the signals is binary: the same description as Figure~\ref{fig:analog} applies.}
    \label{fig:binary}
\end{figure}

\paragraph{Preprocessing.}
The two series of a discrepancy are aligned onto a common timeline
by step-function interpolation. In our data the second series is
sampled at a higher frequency and its timestamps contain those of
the first; we take the timestamps of the second series as the
reference, and for every reference timestamp $t_k$ absent from the
first series we set the interpolated value to the last observed
one.
Step-function interpolation is the natural choice
for monitoring signals that hold their value between measurements,
and avoids creating artificial intermediate values on binary
channels. Discrepancies for which one of the two series is empty
are discarded. After alignment, each discrepancy is represented as
two time-augmented paths in $\mathbb{R}^{3}$:
\[
    X_i^{\mathrm{val}} = (t, x^{1}, x^{2}),
    \qquad
    X_i^{\mathrm{vld}} = (t, v^{1}, v^{2}).
\]
Finally, timestamps and values are normalized to $[0,1]$ using the
minimum and maximum of the \emph{parent signal} (not of the
discrepancy window): this preserves both the relative duration of
each discrepancy and its relative position within the signal. In
particular, discrepancies occurring at system initialization
remain identifiable as such.

\paragraph{Dual-signature representation.}
The two paths are encoded \emph{separately} and their signatures
concatenated, together with the normalized start time of the
discrepancy:
\begin{equation}
    \label{eq:phi_industrial}
    \Phi(d_i)
    = \left(
      t_i^{\mathrm{start}},\;
      \rho\!\left(S^{N}(X_i^{\mathrm{val}})\right),\;
      \rho\!\left(S^{N}(X_i^{\mathrm{vld}})\right)
      \right)
    \in \mathbb{R}^{2 m_{N} + 1},
    \qquad
    m_N = \frac{3^{N+1} - 1}{2}.
\end{equation}
Keeping the two signatures separate preserves the physical
structure of the data: a discrepancy is triggered either by the
values or by the validities, and the non-triggering pair still
carries complementary information about the cause of that discrepancy. Mixing the four
channels into a single path in $\mathbb{R}^{5}$ would entangle
value/validity interactions of no physical meaning while
inflating the dimension.

\paragraph{Resulting pipeline.}
The complete pipeline is summarized in
Figure~\ref{fig:pipeline}: raw discrepancies are aligned and
normalized, encoded by Eq.~\eqref{eq:phi_industrial}, and the
coreset is selected by greedy minimization of
Eq.~\eqref{eq:coreset_problem}, computed exactly by
Eq.~\eqref{eq:breakpoints} with GPU-batched candidate evaluation.
The selected discrepancies are then handed to the
annotation process, and the downstream Transformer classifier (Subsection~\ref{subsec:ind-setup})
is trained on the annotated coreset.

\newpage

\section{Experiments}
\label{sec:experiments}
\subsection{Industrial dataset}

The context and formalization of the industrial dataset are described in
Section~\ref{subsec:ind-inst}; we refer to~\cite{anas} for further details. In
this subsection, we report the dataset's qualitative and quantitative
characteristics, along with the hardware and parameter choices used in our
experiments, before presenting and discussing our results.

\subsubsection{Experimental setup}
\label{subsec:ind-setup}

\paragraph{Data.}
\label{par:data}

The industrial instantiation of Section~\ref{subsec:ind-inst} explicits the data preprocessing protocol.
We here remind that the discrepancies split into two disjoint families according to the nature of
the signal value $x^j$: analog discrepancies, where $x^j$ takes real values,
and binary discrepancies, where $x^j \in \{0,1\}$. Coreset selection is
performed independently within each family, as analog and binary discrepancies
are treated as two separate classification problems. The analog set comprises
$15{,}016$ discrepancies labeled among $9$ fault classes, while the binary set
comprises $5{,}035$ discrepancies labeled among $8$ fault classes. Both sets
exhibit pronounced class imbalance; in the analog set, for instance, the
majority class contains $5{,}964$ discrepancies against only $84$ for the
minority class, a ratio exceeding $70{:}1$.

\begin{table}[htbp]
\centering
\caption{Class distribution for the industrial dataset, by signal family.}
\label{tab:class-dist}
\begin{subtable}[t]{0.48\textwidth}
\centering
\caption{Analog}
\label{tab:analog-class-dist}
\begin{tabular}{lrrr}
\toprule
Class & Train & Test & Total \\
\midrule
Class 1  & 667 & 50  & 717 \\
Class 2  & 1885 & 50  & 1935 \\
Class 3  & 4654 & 50  & 4704 \\
Class 4  & 722 & 50  & 772 \\
Class 6  & 207 & 50  & 257 \\
Class 7  & 5914 & 50  & 5964 \\
Class 8  & 116 & 50  & 166 \\
Class 10  & 34 & 50  & 84 \\
Class 11  & 367 & 50  & 417 \\
\midrule
Total    & 14{,}566 & 450 & 15{,}016 \\
\bottomrule
\end{tabular}
\end{subtable}
\hfill
\begin{subtable}[t]{0.48\textwidth}
\centering
\caption{Binary}
\label{tab:binary-class-dist}
\begin{tabular}{lrrr}
\toprule
Class & Train & Test & Total \\
\midrule
Class 1  & 12 & 20  & 32 \\
Class 2  & 334 & 20  & 354 \\
Class 3  & 44 & 20  & 64 \\
Class 4  & 1111 & 20  & 1131 \\
Class 6  & 417 & 20  & 437 \\
Class 7  & 1512 & 20  & 1532 \\
Class 8  & 479 & 20  & 499 \\
Class 9  & 966 & 20  & 986 \\
\midrule
Total    & 4{,}875 & 160 & 5{,}035 \\
\bottomrule
\end{tabular}
\end{subtable}
\end{table}

We use a train/test split with no dedicated validation set. For each family,
the test set is built by drawing a fixed number of discrepancies uniformly at
random per class, so that it is exactly balanced across classes: $50$
discrepancies per class for the analog set ($450$ discrepancies total across
$9$ classes) and $20$ discrepancies per class for the binary set ($160$
discrepancies total across $8$ classes). The remaining discrepancies form the
training set and retain the original class imbalance. Balancing the test set
ensures that test-time metrics, accuracy in particular, remain informative
despite the imbalance present at training time.

\paragraph{Coreset Budgets.}

We evaluate coreset selection across a grid of budget percentages, computed
globally with respect to the full training set of each family (analog or
binary), rather than per class. This global definition means that, at the
smallest budgets, a method has no guarantee of retaining even one example per
class, making budget selection itself a direct test of whether a method
identifies the relevant classes from the very start, rather than relying on an
enforced coverage floor. We test the grid
$\{0.5, 1, 5, 10, 15, 20, 25, 30, 35, 40, 50, 60, 75, 90\}\%$, with a denser
sampling at low budgets, the regime of primary interest, to characterize how
aggressively the dataset can be reduced before performance degrades, and a
coarser sampling toward higher budgets to confirm convergence toward
full-dataset performance as a sanity check.

\paragraph{Baselines.}

We opt for classical selection methods, namely herding, $k$-means, and facility location. However, there are no existing methods directly applicable to our dataset, so the signature preprocessing step is still applied.
Because herding, $k$-means, and facility location are designed to operate on
point clouds in a fixed-dimensional Euclidean space, and no established
recipe exists in the literature for applying them directly to irregular,
per-channel multivariate time series, we apply all baselines in this
subsection to the same truncated signature representation used by our
method, rather than to the raw discrepancies. This ensures that any
difference in downstream performance reflects the selection criterion
itself rather than a difference in feature space. A method able to operate
directly on raw, non-encoded sequences is examined separately in the UEA
generalization study (Section~\ref{subsec:uea}).

We compare against the following unsupervised, label-agnostic baselines,
none of which use class labels during selection:

\begin{itemize}
    \item \textbf{Random}: uniformly samples $k$ discrepancies from the
    training set without replacement.
    \item \textbf{$k$-means}: clusters the signature representations into
    $k$ clusters using Lloyd's algorithm~\citep{lloyd} with
    $k$-means++ initialization~\citep{kmeans++}, via the
    \texttt{scikit-learn} implementation~\citep{skl} with
    default hyperparameters, and selects the training point closest to each
    centroid as the coreset. Unlike the
    other methods, a new clustering run is performed independently for each
    budget percentage, since $k$-means does not produce a nested sequence of
    coresets across values of $k$.
    \item \textbf{Herding}~\citep{herding, herding_2}: greedily and iteratively selects
    discrepancies to match the running mean of the selected subset to the
    mean of the full training set in signature space, implemented manually
    following the original formulation.
    \item \textbf{Facility location}~\citep{fcltyloc, fcltyloc_2}: greedily selects
    discrepancies to maximize a submodular coverage objective over the
    signature representation, implemented manually following the original
    formulation.
\end{itemize}

For Random, Herding, and Facility location, a single ordered selection is
computed once up to the full training set size ($k = N$), and coresets at
each budget percentage are obtained by truncating this ordered sequence to
the desired prefix length; $k$-means requires an independent run per
budget, as described above.

In addition to these baselines, we evaluate a weighted variant of our own
method, denoted \textbf{SOTS-weighted}. Rather than performing subset selection directly,
SOTS-weighted parameterizes a continuous weighting of the entire training
set on the simplex, via a softmax over learnable logits, and optimizes
these weights with Adam ($100$ projections, $5000$ iterations, learning
rate $0.05$) to minimize the sliced Wasserstein distance between the
resulting weighted empirical measure and the full training set's empirical
measure, both in signature space. The coreset at budget $k$ is then
obtained by discarding the optimized weights and retaining only the
indices of the $k$ highest-weighted points, following the same single-run,
prefix-truncation pattern as Herding and Facility location.

\paragraph{SOTS Configuration.}

Signatures are truncated at level $N=2$ (an ablation study of the impact of the signature order is conducted in Section~\ref{subsec:uea}); the path construction from the two
independent per-signal grids and the level-wise rescaling applied to correct
scale imbalance are detailed in Section~\ref{subsec:ind-inst}. Sliced Wasserstein
distances are computed with $p=2$ and $L=100$ projections throughout, a
value chosen given the $27$-dimensional signature representation, beyond
which additional projections did not meaningfully change the estimated
distance. We implement four projection strategies: standard uniform
projections, energy-based sliced Wasserstein
(EBSW,~\citep{ebsw}), and two random-path projection variants,
RPSW and IWRPSW~\citep{rpsw}, which replace uniform slicing
directions with directions sampled along random paths between dataset
points, with (IWRPSW) or without (RPSW) an additional importance weighting
of the resulting per-direction distances. Our main results use the
uniform variant; we compare all four
projection strategies directly in Section~\ref{subsubsec:results1}, restricted
to the binary dataset and budgets between $5\%$ and $40\%$, the regime in which the choice of
projection strategy still measurably affects downstream performance before
all variants converge to comparable behavior at larger budgets.

\paragraph{Downstream Classifier.}

Coreset quality is evaluated by training the Transformer architecture
of~\citep{anas} from scratch on each selected coreset. The
architecture is an encoder-only Transformer following Vaswani et
al.~\citep{attention}, adapted for irregular multivariate time
series classification as in the SEFT framework~\citep{seft}: each timestamp is represented as a vector
of channel values concatenated with a binary measurement indicator and a
positional time embedding, before being processed by $1$ Transformer
encoder layer with hidden dimension $64$ and $4$ attention heads,
mean-aggregated, and passed to a fully connected classification layer.
Unlike the binary false-positive detection task
of~\citep{anas}, we adapt the classification head to output a
softmax over the $9$ (analog) or $8$ (binary) fault classes of the
corresponding track. Classifier training uses the authors' own
preprocessing pipeline (downsampling and positional time
embedding, as described above) applied directly to the raw discrepancies;
the signature representation used for coreset selection is not used as
classifier input, and serves only to determine which discrepancies are
included in the training coreset. We use the authors' PyTorch
implementation, training with Adam, learning rate $0.001$, batch size
$64$, and $1000$ warmup steps, for a fixed $150$ epochs per (method,
budget) configuration. Unlike~\citep{anas}, we do not use
validation-based early stopping, since our dataset has no dedicated validation split.

\paragraph{Evaluation Metrics.}

We report three metrics: accuracy, macro-averaged F1, and macro-averaged
Area Under the Precision-Recall Curve (AUPRC). Accuracy is
reported as a standard classification metric; because our test sets are
constructed to be exactly balanced across classes, accuracy here is equivalent to macro-averaged recall, and is
therefore not inflated by majority-class performance the way it would be
under the natural class distribution, where a model predicting only the
majority class could otherwise achieve high accuracy while ignoring rare
fault classes entirely. F1-macro is reported as our primary metric, as it
is unweighted, so each class contributes equally to the score regardless
of its frequency in training. AUPRC is retained from~\citep{anas},
which uses it as one of its three reported metrics, and is included here
for comparability with that prior evaluation.

\paragraph{Implementation and hardware.}
All coreset selection methods (SOTS, Herding, Facility location, $k$-means) are implemented in Python, with
PyTorch used for the GPU-parallelized components of the greedy selection
procedures. Path signatures are computed using a custom implementation.
Experiments are run on a single NVIDIA A100 80GB PCIe
GPU, using Python 3.12.7 and PyTorch 2.11.0 (CUDA 13.0). Per-method runtime is reported alongside the corresponding results.

\subsubsection{Results}
\label{subsubsec:results1}

\begin{figure}[H]
            \centering
            \includegraphics[width=0.75\linewidth]{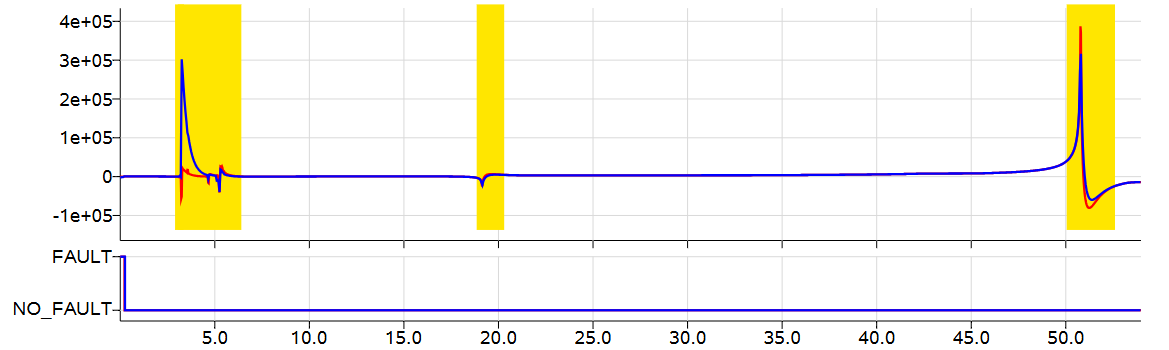}
            \begin{table}[H]
                \centering
                \resizebox{\textwidth}{!}{%
                \begin{tabular}{c|ccccccccccccccccccc}
                \toprule
                ID & $d_0$ & $d_1$ & $d_2$ & $d_3$ & $d_4$ & $d_5$ & $d_6$ & $d_7$ & $d_8$ & $d_9$ & $d_{10}$ & $d_{11}$ & $d_{12}$ & $d_{13}$ & $d_{14}$ & $d_{15}$ & $d_{16}$ & $d_{17}$ & $d_{18}$ \\
                \midrule
                1 & 0.05 & 1.00 & 0.08 & -0.00 & -0.00 & 0.01 & -0.00 & -0.01 & 0.00 & -0.09 & -0.32 & 0.01 & 0.26 & -0.80 & 1.00 & 0.08 & 0.00 & 0.00 & 0.01 \\
                2 & 0.34 & 1.00 & 0.04 & 0.17 & 0.16 & 0.00 & 0.01 & 0.01 & 0.00 & -0.30 & 0.36 & 0.00 & -0.52 & -0.27 & 1.00 & 0.04 & 0.00 & 0.00 & 0.00 \\
                3 & 0.91 & 1.00 & 0.06 & -0.11 & -0.11 & 0.00 & -0.01 & -0.01 & -0.00 & -0.31 & 0.18 & -0.00 & -0.36 & -0.22 & 1.00 & 0.06 & 0.00 & 0.00 & 0.00 \\
                \bottomrule
                \end{tabular}%
                }
            \end{table}
            \caption{Example of analog discrepancies and their signature vectors}
            \label{image4}
        \end{figure}

Before examining metrics for each coreset budget, we first illustrate what
a signature vector concretely looks like. Figure/Table~\ref{image4} shows
three discrepancies from the analog signal family, together with their
signature vectors; coordinates that are uniformly zero are omitted for
readability. The first coordinate corresponds to the discrepancy's
normalized start time. The order-1 terms, which capture the offset
between each channel's start and end value, remain zero throughout for the first discrepancy ($d_{3}$ and $d_{4}$), since
by construction the signal returns to its initial value by the end of the
discrepancy window. The order-2 terms carry the most informative content
in this example, in particular $d_{10}$ and $d_{12}$, which encode the
L\'evy areas between the two compared signals. The remaining coordinates,
from $d_{14}$ onward, correspond to the signature of the validity grid;
since both signals remain valid throughout this example, this portion of
the vector is almost entirely zero and carries little discriminative
information here. Despite being drawn from three distinct discrepancies,
the resulting signature vectors share many close or matching coordinate
values, consistent with the fact that all three belong to the same class
(label 10).

\begin{figure}[H]
    \centering
    \begin{minipage}{0.48\textwidth}
        \centering
        \includegraphics[width=1\linewidth]{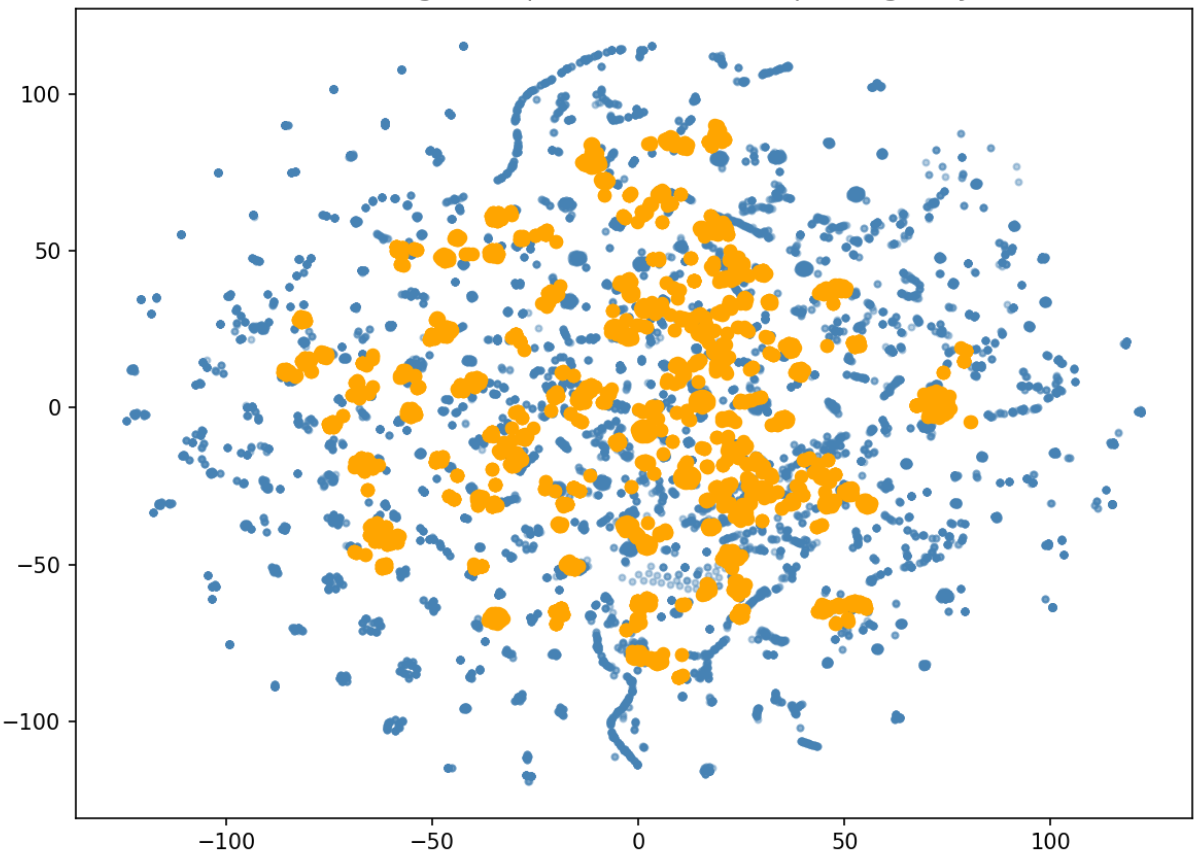}
        \caption{Weighted coreset distribution ($20\%$ budget, analog dataset)}
        \label{fig:weighted-tsne}
    \end{minipage}
    \hfill
    \begin{minipage}{0.48\textwidth}
        \centering
        \includegraphics[width=1\linewidth]{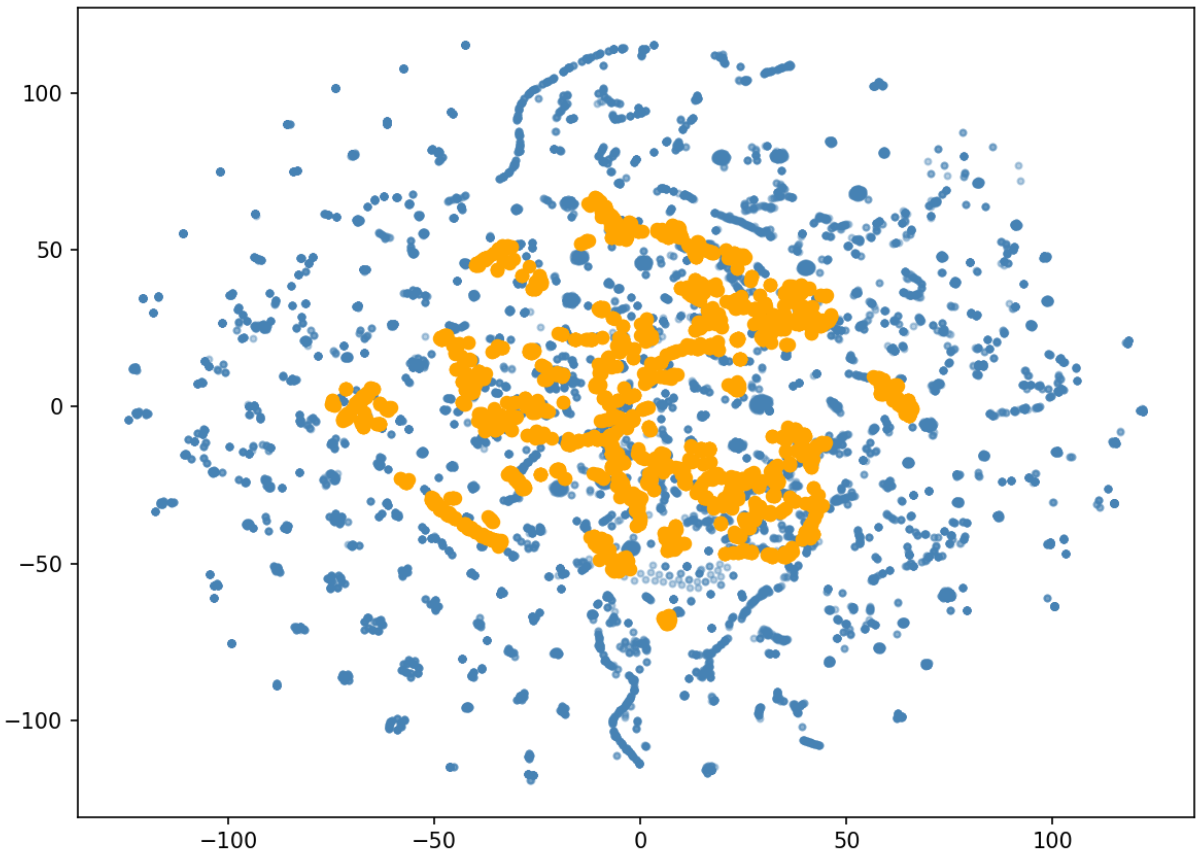}
        \caption{Greedy coreset distribution ($20\%$ budget, analog dataset)}
        \label{fig:greedy-tsne}
    \end{minipage}
\end{figure}

To complement the signature vector illustration above, we visualize how a
candidate coreset approximates the distribution of its parent dataset.
Figures~\ref{fig:weighted-tsne} and~\ref{fig:greedy-tsne} show a
two-dimensional t-SNE projection (perplexity $=30$) of the analog
signature representation, with a $20\%$-budget coreset selected by
SOTS-weighted (left) and SOTS-greedy (right) highlighted in orange. As
this is a two-dimensional projection of the original $27$-dimensional
signature space, distances and apparent cluster structure in this
visualization should be interpreted with caution; it is used here only to
illustrate coverage qualitatively, not as a quantitative comparison. The
weighted coreset appears to reach further into the periphery of the point
cloud, covering more of the sparse, outlying regions, while the greedy
coreset concentrates more densely around the central region of the
embedding. Despite this apparently broader coverage, SOTS-weighted
substantially underperforms SOTS-greedy in downstream classification
accuracy (Table~\ref{tab:results_industrial}). This illustrates a central
point: visually covering the full distribution is not the same as selecting discriminative, class-relevant instances,
thus a broad visual coverage in this projection should not be read as a
proxy for downstream quality.

\begin{figure}[H]
    \centering
    \begin{subfigure}{0.24\textwidth}
        \centering
        \includegraphics[width=\textwidth]{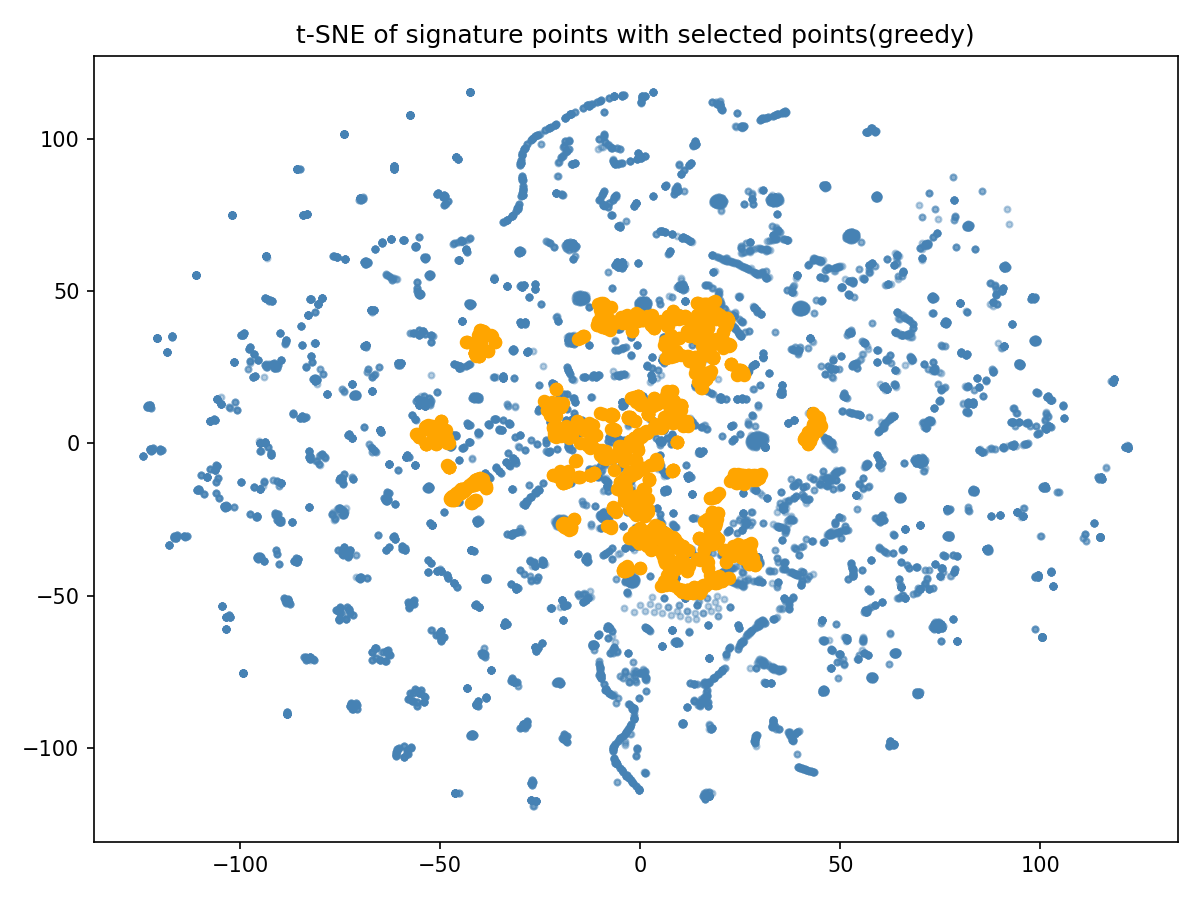}
        \label{fig:sub1}
    \end{subfigure}
    \hfill
    \begin{subfigure}{0.24\textwidth}
        \centering
        \includegraphics[width=\textwidth]{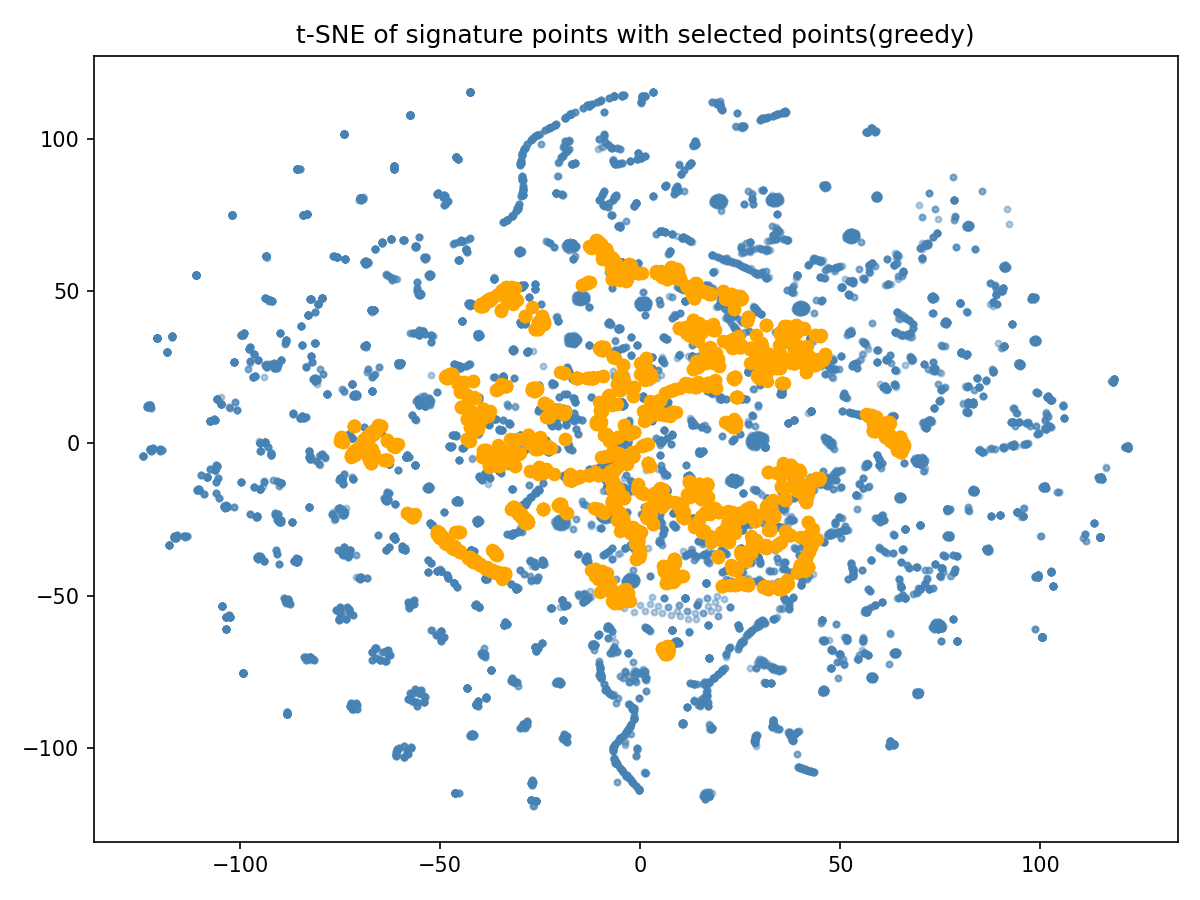}
        \label{fig:sub2}
    \end{subfigure}
    \hfill
    \begin{subfigure}{0.24\textwidth}
        \centering
        \includegraphics[width=\textwidth]{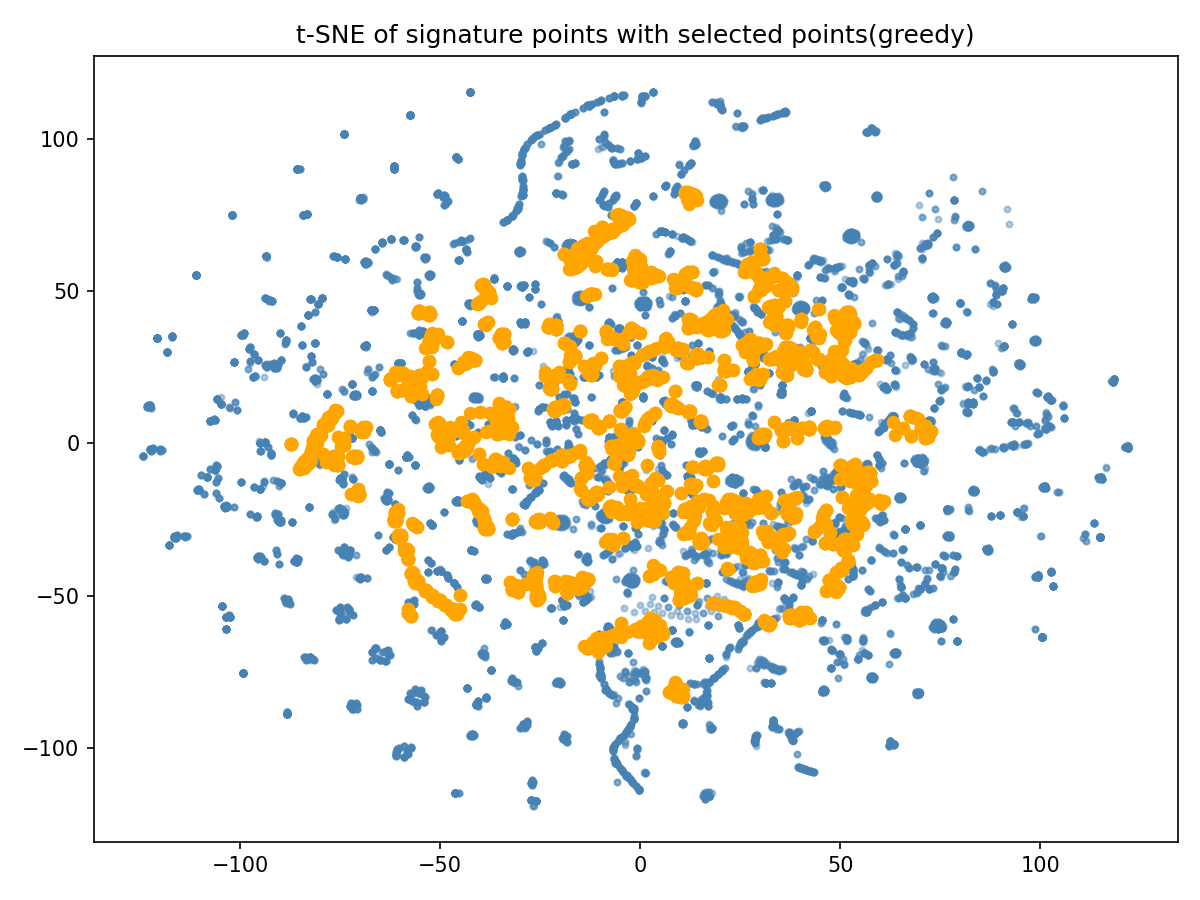}
        \label{fig:sub3}
    \end{subfigure}
    \hfill
    \begin{subfigure}{0.24\textwidth}
        \centering
        \includegraphics[width=\textwidth]{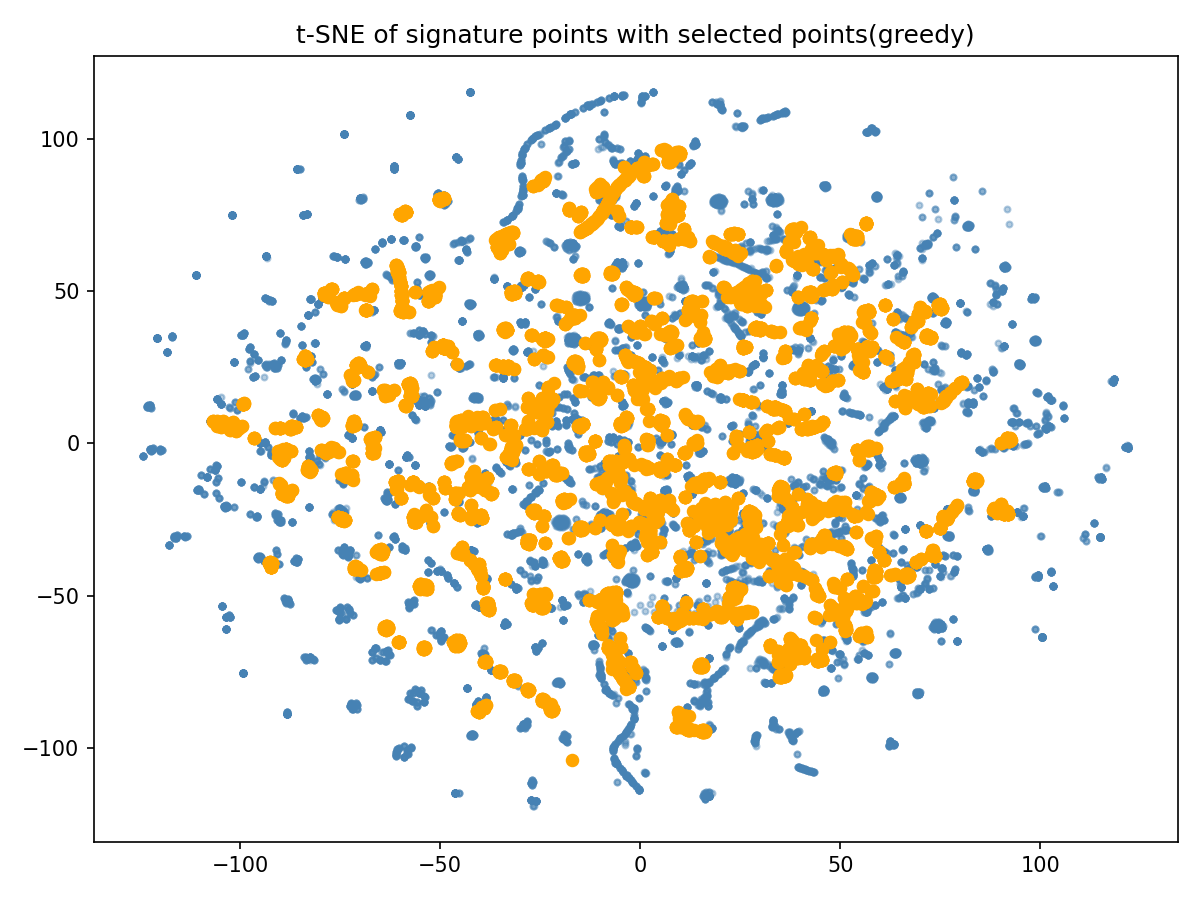}
        \label{fig:sub4}
    \end{subfigure}
    \caption{Evolution of SOTS-greedy coreset coverage (10-20-30-50\%)}
    \label{fig:combined}
\end{figure}

We now examine how the SOTS-greedy coreset's
coverage of the signature space evolves as the budget increases.
Figure~\ref{fig:combined} shows the same t-SNE projection
(analog dataset, perplexity $=30$) with the SOTS-greedy coreset highlighted
at four increasing budgets, $10\%$, $20\%$, $30\%$, and $50\%$, allowing
the progressive densification of the selected subset to be observed
directly.

Table~\ref{tab:results_industrial} reports results on the analog dataset.
SOTS-Greedy is the strongest method overall, ranking first in $57\%$ of
all (budget, metric) cells and within the top two in $86\%$ of them, most
markedly from $15\%$ budget onward, where it is first or second across
nearly every remaining budget and metric. At the extreme low end
($0.5\%$), SOTS-Greedy already achieves the best accuracy and F1-macro of
any method. Between $1\%$ and $10\%$, $k$-means and Facility location
briefly overtake it on accuracy and F1-macro, though SOTS-Greedy remains
competitive on AUPRC throughout this range; we do not consider this
narrow dip practically significant, as absolute performance at these
budgets remains far below full-dataset performance regardless of method.
Herding and SOTS-weighted are essentially never competitive on this
track, never appearing in the top two across any (budget, metric) cell.

\begin{table}[H]
\resizebox{\textwidth}{!}{%
\begin{tabular}{p{2.5cm}lcccccccccccccc}
\toprule
\centering \textbf{Method} & \textbf{Metric}
& \textbf{0.5\%} & \textbf{1\%} & \textbf{5\%}
& \textbf{10\%} & \textbf{15\%} & \textbf{20\%} & \textbf{25\%}
& \textbf{30\%} & \textbf{35\%} & \textbf{40\%} & \textbf{50\%}
& \textbf{60\%} & \textbf{75\%}
& \textbf{90\%}  \\
\midrule

\multirow{3}{2.5cm}{\centering Random}
& Accuracy & 46.44 & 52.89 & 70.00 & 80.67 & 83.56 & 84.00 & 85.33 & 85.78 & 88.00 & 89.56 & 91.78 & 91.33 & {\color{blue} 94.00} & 93.33 \\
& F1 Macro & 40.63 & 48.90 & 69.95 & 80.42 & 83.69 & 83.85 & 85.25 & 85.85 & 88.09 & 89.62 & 91.65 & 91.25 & {\color{blue} 93.90} & 93.25 \\
& AUPRC    & 42.54 & 48.08 & 75.99 & 86.62 & 87.15 & 88.76 & 89.10 & 91.51 & 92.69 & 94.26 & 95.41 & 95.92 & {\color{blue} 97.66} & 97.42 \\
\midrule

\multirow{3}{2.5cm}{\centering K-Means + Sig.}
& Accuracy & {\color{blue} 48.22} &{\color{blue}  60.89} & \textbf{79.33} & \textbf{83.78} &{\color{blue} 86.67} & {\color{blue} 86.67} & {\color{blue} 86.44} & 88.22 & \textbf{94.00} & \textbf{93.78} & \textbf{94.44} & \textbf{94.44} & 93.56 & {\color{blue} 94.22} \\
& F1 Macro & {\color{blue} 44.86} & {\color{blue} 60.69} & \textbf{79.23} & \textbf{84.20} & {\color{blue} 86.84} & {\color{blue} 86.78} & {\color{blue} 86.64} & 88.51 & \textbf{93.99} & \textbf{93.76} & \textbf{94.44} & \textbf{94.42} & 93.49 & {\color{blue} 94.18} \\
& AUPRC    & {\color{blue} 47.35} & \textbf{60.70} & {\color{blue} 81.25} & {\color{blue} 86.34} & 88.54 & 89.11 & 89.04 & 92.78 & \textbf{96.24} & {\color{blue} 96.41} &{\color{blue} 96.96} & {\color{blue} 97.44} & 96.81 & {\color{blue} 97.62} \\
\midrule

\multirow{3}{2.5cm}{\centering Herding + Sig.}
& Accuracy & 34.67 & 42.00 & 55.78 & 60.44 & 68.44 & 74.22 & 77.33 & 79.33 & 78.89 & 81.11 & 85.11 & 87.78 & 90.44 & 93.33 \\
& F1 Macro & 34.00 & 38.74 & 54.93 & 60.40 & 68.79 & 74.43 & 77.48 & 79.51 & 78.93 & 81.34 & 85.48 & 88.11 & 90.52 & 93.27 \\
& AUPRC    & 38.77 & 44.14 & 55.07 & 64.12 & 73.61 & 77.97 & 82.08 & 86.67 & 84.38 & 87.92 & 92.22 & 92.77 & 96.74 & 97.18 \\
\midrule

\multirow{3}{2.5cm}{\centering Facility Loc. + Sig.}
& Accuracy & 47.11 & \textbf{61.33} & {\color{blue} 78.67} & {\color{blue} 82.67} & 84.67 & 85.78 & 85.56 & {\color{blue} 90.44} & 89.78 & 88.22 & 90.00 & 90.00 & 91.56 & 92.22 \\
& F1 Macro & 43.31 & \textbf{61.71} & 78.56 & {\color{blue} 83.05} & 84.99 & 86.24 & 86.11 & {\color{blue} 90.53} & 89.88 & 88.54 & 90.22 & 90.03 & 91.58 & 92.14 \\
& AUPRC    & \textbf{47.69} & {\color{blue} 60.18} & 81.21 & 85.26 & {\color{blue} 88.76} & {\color{blue} 89.48} & {\color{blue} 90.02} & {\color{blue} 94.41} & 94.97 & 95.19 & 96.05 & 95.44 & 96.65 & 97.12 \\
\midrule

\multirow{3}{2.5cm}{\centering SOTS-Weighted}
& Accuracy & 30.22 & 42.00 & 64.44 & 66.67 & 71.11 & 76.22 & 75.78 & 76.22 & 78.67 & 79.11 & 81.33 & 84.44 & 86.22 & 90.89 \\
& F1 Macro & 26.80 & 38.44 & 63.12 & 66.59 & 70.61 & 76.24 & 76.08 & 76.51 & 79.03 & 78.96 & 81.14 & 84.38 & 86.35 & 90.80 \\
& AUPRC    & 29.17 & 41.41 & 64.66 & 71.75 & 74.81 & 81.99 & 82.97 & 82.59 & 84.38 & 85.01 & 88.47 & 90.29 & 93.29 & 96.03 \\
\midrule

\multirow{3}{2.5cm}{\centering \textbf{SOTS-Greedy (ours)}}
& Accuracy & \textbf{48.67} & 59.11 & 78.22 & 82.00 & \textbf{86.89 }& \textbf{87.33} & \textbf{88.67} & \textbf{91.56} & {\color{blue} 91.78} & {\color{blue} 93.56} & {\color{blue} 94.00} & {\color{blue} 94.00} & \textbf{94.44} & \textbf{94.89} \\
& F1 Macro & \textbf{44.97} & 57.65 & {\color{blue} 78.62} & 82.28 & \textbf{86.96} & \textbf{87.45} & \textbf{88.75} & \textbf{91.63} & {\color{blue} 91.88} & {\color{blue} 93.55} & {\color{blue} 93.96} & {\color{blue} 93.94} & \textbf{94.38} & \textbf{94.83} \\
& AUPRC    & 44.79 & 60.65 & \textbf{86.07} & \textbf{88.67} & \textbf{92.10} & \textbf{90.45} & \textbf{94.06} & \textbf{95.12} & {\color{blue} 95.29} & \textbf{97.51} & \textbf{97.53} & \textbf{97.94} & \textbf{96.99} & \textbf{98.29} \\
\bottomrule
\end{tabular}%
}
\centering
\caption{
    Classification performance on the industrial analog dataset
    for each coreset selection method across budget ratios (all metrics in \%).
    \textbf{Bold} denotes the best value and
    {\color{blue} blue} denotes the second best at each ratio.
}
\label{tab:results_industrial}
\end{table}

Table~\ref{tab:results_industrial_binary} reports the corresponding
results on the binary dataset, and the picture changes substantially.
SOTS-Greedy falls to the fourth most competitive method by top-2 rate
($33\%$), behind $k$-means ($67\%$), Facility location ($55\%$), and even
Random sampling ($43\%$), which alone wins outright at $0.5\%$, $1\%$,
and $5\%$ budgets on accuracy and F1-macro. Binary discrepancies take
only $\{0,1\}$ values, with far less continuous geometric variation than
their analog counterparts; we hypothesize that this comparative
redundancy across the resulting signature vectors makes it harder for a
distributional-matching selection criterion to distinguish informative
from uninformative instances at small budgets, though we do not verify
this mechanism directly and treat it as an open question. As on the analog dataset, Herding and
SOTS-weighted remain uncompetitive throughout.

\begin{table}[H]
\resizebox{\textwidth}{!}{%
\begin{tabular}{p{2.5cm}lcccccccccccccc}
\toprule
\centering \textbf{Method} & \textbf{Metric}
& \textbf{0.5\%} & \textbf{1\%} & \textbf{5\%}
& \textbf{10\%} & \textbf{15\%} & \textbf{20\%} & \textbf{25\%}
& \textbf{30\%} & \textbf{35\%} & \textbf{40\%} & \textbf{50\%}
& \textbf{60\%} & \textbf{75\%}
& \textbf{90\%}  \\
\midrule

\multirow{3}{2.5cm}{\centering Random}
& Accuracy & \textbf{66.88} & \textbf{74.38} & \textbf{90.63} & 81.88 & {\color{blue} 91.25} & 92.50 & {\color{blue} 93.13} & 92.50 & 94.38 & 95.00 & {\color{blue} 95.63} & {\color{blue} 96.88} & {\color{blue} 96.88} & \textbf{96.88} \\
& F1 Macro & \textbf{62.54} & \textbf{70.70} & \textbf{90.81} & 78.15 & 91.45 & 92.56 & 93.16 & 92.53 & 94.37 & 95.01 & {\color{blue} 95.61} & {\color{blue} 96.91} & {\color{blue} 96.90} & \textbf{96.90} \\
& AUPRC    & \textbf{78.75} & \textbf{79.82} & \textbf{94.32} & 90.51 & 93.01 & 95.63 & \textbf{97.97} & 96.23 & 97.42 & {\color{blue} 98.23} & 98.54 & 98.93 & 98.81 & 99.03 \\
\midrule

\multirow{3}{2.5cm}{\centering K-Means + Sig.}
& Accuracy & 30.63 & {\color{blue} 61.88} & 82.50 & {\color{blue} 89.38} & 90.00 & \textbf{94.38} & {\color{blue} 93.13} & \textbf{95.63} & \textbf{96.88} & \textbf{96.88} & 95.00 & \textbf{97.50} & \textbf{97.50} & \textbf{96.88} \\
& F1 Macro & 23.66 & {\color{blue} 60.52} & 82.39 & {\color{blue} 89.27} & 90.19 & \textbf{94.39} & 93.12 & \textbf{95.69} & \textbf{96.90} & \textbf{96.87} & 95.05 &\textbf{ 97.53} & \textbf{97.52} & \textbf{96.90} \\
& AUPRC    & 37.17 & 60.92 & 88.28 & \textbf{96.27} & \textbf{96.47} & {\color{blue} 96.96} & 97.27 & \textbf{99.36} & \textbf{99.47} & \textbf{99.62} & {\color{blue} 98.78} & \textbf{99.67} & \textbf{99.46} & \textbf{99.51} \\
\midrule

\multirow{3}{2.5cm}{\centering Herding + Sig.}
& Accuracy & 30.63 & 30.63 & 51.25 & 76.88 & 81.25 & 85.00 & 86.25 & 85.63 & 85.63 & 85.63 & 87.50 & 88.13 & 89.38 & 90.63 \\
& F1 Macro & 21.41 & 25.38 & 49.60 & 76.22 & 80.65 & 84.28 & 85.34 & 84.46 & 84.87 & 84.96 & 86.63 & 87.26 & 88.93 & 90.82 \\
& AUPRC    & {\color{blue} 39.33} & 34.76 & 55.80 & 82.67 & 85.48 & 91.52 & 91.71 & 93.43 & 94.95 & 94.93 & 94.97 & 96.07 & 97.30 & 97.99 \\
\midrule

\multirow{3}{2.5cm}{\centering Facility Loc. + Sig.}
& Accuracy & {\color{blue} 42.50} & 59.38 & {\color{blue} 85.00} & \textbf{91.88} & \textbf{92.50 }& {\color{blue} 93.13} & {\color{blue} 93.13} & 91.25 & {\color{blue} 96.25} & 95.00 & 93.75 & 96.25 & \textbf{97.50} & \textbf{96.88} \\
& F1 Macro & {\color{blue} 32.87} & 55.31 & {\color{blue} 84.81} & \textbf{91.96} & \textbf{92.62} & {\color{blue} 93.11} & {\color{blue} 93.17} & 91.41 & {\color{blue} 96.25} & 95.06 & 93.78 & 96.27 & \textbf{97.52} & \textbf{96.90} \\
& AUPRC    & 38.56 & {\color{blue} 62.52} & {\color{blue} 91.81} & {\color{blue} 94.83} & 95.20 & \textbf{97.15} & 96.38 & 97.94 & {\color{blue} 98.47} & 97.84 & 98.24 & {\color{blue} 99.00} & {\color{blue} 99.28} & {\color{blue} 99.53} \\
\midrule

\multirow{3}{2.5cm}{\centering SOTS-Weighted}
& Accuracy & 21.25 & 29.38 & 41.88 & 52.50 & 57.50 & 74.38 & 77.50 & 78.13 & 80.00 & 82.50 & 82.50 & 88.13 & 89.38 & {\color{blue} 91.25} \\
& F1 Macro & 14.65 & 18.96 & 33.92 & 49.57 & 53.47 & 74.23 & 77.05 & 78.20 & 79.84 & 82.77 & 82.24 & 88.25 & 89.33 & {\color{blue} 91.34} \\
& AUPRC    & 31.87 & 27.85 & 48.26 & 55.68 & 61.32 & 80.87 & 82.10 & 83.87 & 85.10 & 86.52 & 89.53 & 92.14 & 94.13 & 95.77 \\
\midrule

\multirow{3}{2.5cm}{\centering \textbf{SOTS-Greedy (ours)}}
& Accuracy & 34.38 & 34.38 & 80.00 & 81.88 &\textbf{ 92.50} & 91.25 & \textbf{94.38} & {\color{blue} 94.38} & 95.63 &{\color{blue} 96.25} & \textbf{96.88} & {\color{blue} 96.88} & 96.25 & \textbf{96.88} \\
& F1 Macro & 26.36 & 26.87 & 79.83 & 81.79 & {\color{blue} 92.49} & 91.12 & \textbf{94.39} & {\color{blue} 94.30} & 95.60 & {\color{blue} 96.27} & \textbf{96.90} & 96.90 & 96.28 & \textbf{96.90} \\
& AUPRC    & 32.06 & 44.77 & 87.16 & 92.04 & {\color{blue} 96.85} & 96.44 & {\color{blue} 97.31} & {\color{blue} 98.21} & 97.67 & 97.66 & \textbf{98.93} & 98.28 & 98.28 & 98.84 \\

\bottomrule
\end{tabular}%
}
\centering
\caption{
    Classification performance on the industrial binary dataset
    for each coreset selection method across budget ratios (all metrics in \%).
    \textbf{Bold} denotes the best value and
    {\color{blue} blue} denotes the second best at each ratio.
}
\label{tab:results_industrial_binary}
\end{table}

\paragraph{Effect of the projection strategy.}
Table~\ref{tab:results_projection} compares the uniform projection
strategy used throughout our main results against the three adaptive
variants (EBSW, RPSW, IWRPSW) for SOTS-Greedy on the binary track,
restricted to budgets between $5\%$ and $40\%$. Uniform projection wins
outright in only $1/24$ cells and is strictly beaten by at least one
adaptive variant in $22/24$ ($92\%$) of them. RPSW and IWRPSW are the two
strongest variants and closely matched overall, each winning $8/24$
cells outright; RPSW is markedly stronger at the lowest budget shown
($5\%$, winning all three metrics) and yields the better average AUPRC
improvement over uniform ($+0.13$ vs. $-1.12$ for IWRPSW), while IWRPSW
edges ahead at several mid-range budgets on accuracy and F1-macro. EBSW
improves least over uniform and is, on average, the only variant to
underperform it on both AUPRC and overall win rate. Using each method's
best available projection at each budget, SOTS-Greedy is closely
competitive with $k$-means and Facility location from
Table~\ref{tab:results_industrial_binary} over this same budget range,
winning or tying $11/24$ cells, particularly on accuracy and F1-macro
between $15\%$ and $35\%$ budget, though $k$-means remains ahead on
AUPRC from $30\%$ budget onward even against SOTS-Greedy's best
projection. This gain in performance comes at a real computational cost:
selecting a $40\%$ coreset takes $2.17$h and $2.18$h with RPSW and
IWRPSW respectively, roughly $4.7\times$ the $0.46$h required by uniform
projection, since both variants must select projection directions
specific to the points already chosen at each greedy step. EBSW incurs a
more modest overhead ($0.63$h, $1.4\times$ uniform). We retain uniform
projection throughout our main results as the better
performance-per-compute trade-off.\\

\begin{table}[h]
\resizebox{\textwidth}{!}{%
\begin{tabular}{p{2cm}p{2cm}lcccccccc}
\toprule
\centering \textbf{Method} & \centering \textbf{Time (h)} & \textbf{Metric} 
& \textbf{5\%}
& \textbf{10\%} & \textbf{15\%} & \textbf{20\%} & \textbf{25\%}
& \textbf{30\%} & \textbf{35\%} & \textbf{40\%}  \\
\midrule

\multirow{3}{2cm}{\centering Uniform.}
& \multirow{3}{2cm}{\centering 0.46}
& Accuracy & 80.00 & 81.88 & 92.50 & 91.25 & 94.38 & 94.38 & 95.63 & \textbf{96.25} \\
& & F1 Macro & 79.83 & 81.79 & 92.49 & 91.12 & 94.39 & 94.30 & 95.60 & 96.27 \\
& & AUPRC    & 87.16 & 92.04 & \textbf{96.85} & 96.44 & 97.31 & 98.21 & 97.67 & 97.66\\
\midrule

\multirow{3}{2cm}{\centering EBSW.}
& \multirow{3}{2cm}{\centering 0.63}
& Accuracy & 76.25 & 85.63 & \textbf{93.13} & 92.50 & 93.75 & 95.00 & 96.25 & \textbf{96.25}\\
& & F1 Macro & 76.15 & 85.42 & \textbf{93.30} & 92.55 & 93.83 & 95.03 & 96.28 & \textbf{96.28} \\
& & AUPRC    & 79.76 & 91.09 & 96.69 & \textbf{96.93} & \textbf{97.48} & 97.99 & 97.20 & 97.50 \\
\midrule

\multirow{3}{2cm}{\centering RPSW.}
& \multirow{3}{2cm}{\centering 2.17}
& Accuracy & \textbf{84.38} & 85.63 & 88.13 & \textbf{94.38} & 93.13 & 95.00  & \textbf{96.88} & \textbf{96.25} \\
& & F1 Macro & \textbf{84.13} & 85.55 & 87.75 & \textbf{94.38} & 93.17 & 95.02 & \textbf{96.90} & \textbf{96.28} \\
& & AUPRC    & \textbf{91.17} & 92.69 & 93.24 & 96.40 & 95.76 & 98.94 & \textbf{97.85} & 98.32 \\
\midrule

\multirow{3}{2cm}{\centering IWRPSW.}
& \multirow{3}{2cm}{\centering 2.18}
& Accuracy & 76.88 & \textbf{87.50} & 92.50 & 93.75 & \textbf{95.63} & \textbf{96.88} & 95.63 & \textbf{96.25} \\
& & F1 Macro & 73.81 & \textbf{87.34} & 92.58 & 93.87 & \textbf{95.65} & \textbf{96.87} & 95.66 & \textbf{96.28} \\
& & AUPRC    & 81.58 & \textbf{92.92} & 93.82 & 96.76 & 96.64 & \textbf{96.93} & 97.16 & \textbf{98.58} \\

\bottomrule
\end{tabular}%
}
\centering
\caption{
    Classification performance on the industrial binary dataset
    for SOTS-greedy coreset selection method across budget ratios and projection methods (all metrics in \%).
    \textbf{Bold} denotes the best value.
}
\label{tab:results_projection}
\end{table}

\begin{figure}[H]
    \centering
    \includegraphics[width=1\linewidth]{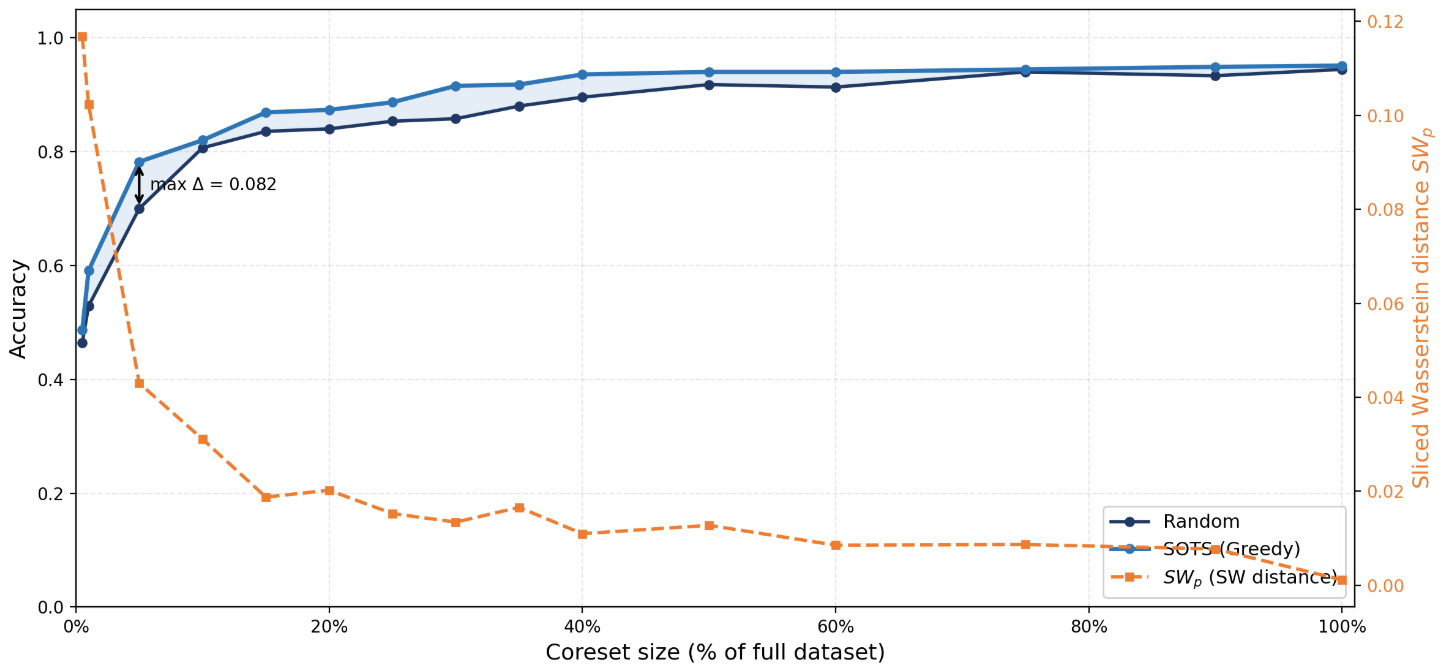}
    \caption{Evolution of downstream classification accuracy and sliced
    Wasserstein distance $SW_p$ between the coreset and the
    full training set, as a function of coreset budget, for
    SOTS-Greedy and Random sampling on the industrial
    analog dataset.}
    \label{fig:sw-perf}
\end{figure}

To empirically validate our initial hypothesis (Eq.~\ref{eq:hypothesis2}),
we compare the evolution of downstream classification performance for the
SOTS-Greedy coreset against the evolution of the sliced Wasserstein
distance between the SOTS-Greedy coreset and the full training set,
shown in Figure~\ref{fig:sw-perf}; Random sampling is included as a
performance baseline. The two quantities track
each other closely across the budget range: sliced Wasserstein distance
falls sharply up to roughly $20$–$30\%$ budget and continues to decrease
more gradually thereafter, while accuracy for both methods rises
correspondingly and both curves flatten together at approximately
$40\%$ budget. SOTS-Greedy's advantage over Random is largest in the
low-budget regime, peaking at $\max \Delta = 0.082$ accuracy near $5\%$
budget, exactly where sliced Wasserstein distance is also falling most
sharply, direct empirical support for the hypothesis precisely where a
coreset method matters most in practice. The coupling between the two
quantities is visible not only in this overall trend but locally: at
$35\%$, sliced Wasserstein distance
increases slightly rather than continuing to fall, and accuracy
correspondingly stalls or dips slightly at the same budget, rather than
continuing its steady improvement. This local co-movement is,
if anything, stronger evidence for the approximate equivalence in
Eq.~\ref{eq:hypothesis2} than the global trend alone, since it would be
coincidental for an unrelated quantity to move in step with performance
at this resolution. Both curves converge as budget approaches
$100\%$, as expected, since the coreset then coincides with the full
training set by construction.

\newpage

\subsection{UEA generalization study}
\label{subsec:uea}

Because SOTS encodes discrepancies through truncated path signatures, it is
inherently sensitive to the geometric structure and shape traced out by a
sample's trajectory, rather than to properties such as raw amplitude or
frequency content considered independently of trajectory shape
(Section~\ref{sec:methods}). We therefore expect SOTS to be most
effective on tasks where class-discriminative information is itself
geometric, as is the case for our industrial discrepancies, and less
effective when discriminative information lies elsewhere. To test this
hypothesis beyond our industrial setting, and to examine the effect of
signature truncation order, we evaluate SOTS on three datasets from the
UEA multivariate time series classification
archive~\citep{uea}.

\subsubsection{Experimental setup}

\paragraph{Data.}

\textbf{Libras} and \textbf{PenDigits}
are both trajectory-based datasets in which each sample is a
two-dimensional curve traced over time: a hand movement for Libras ($15$
classes of Brazilian Sign Language gestures, $2$ channels, series length
$45$, $180$ training and $180$ test instances) and a pen-tip trajectory for
PenDigits ($10$ handwritten digit classes, $2$ channels, series length
$8$, $7{,}494$ training and $3{,}498$ test instances). In both cases the
class label is determined directly by the shape of the traced curve,
making them a direct match for the geometric information captured by path
signatures. \textbf{EthanolConcentration} is
included as a contrasting case: each sample is a raw near-infrared
spectrum of a water-ethanol solution at one of four concentration levels
($3$ channels, series length $1{,}751$, $261$ training and $263$ test
instances), where discriminative information lies in the spectral response
at specific frequencies rather than in any geometric trajectory shape.
Unlike our industrial dataset, all three datasets are used in their
native, regularly sampled, with no irregularity
introduced; this generalization study therefore also tests SOTS outside
the irregular-sampling regime it was designed for. We use the canonical
UEA train/test splits for all three datasets without modification.

\paragraph{Coreset Budgets.}

As in the industrial dataset (Section~\ref{subsec:ind-setup}), we evaluate coreset
selection across the same grid of budget percentages, computed globally
with respect to each dataset's training set. Absolute
coreset sizes are obtained as $k = \max(1, \mathrm{round}(p \cdot n / 100))$,
where $p$ is the budget percentage and $n$ the training set size; this
floor guarantees a non-empty coreset at every budget but is not active for
any (dataset, budget) pair in this grid, as the smallest resulting value,
$1$ instance for Libras and EthanolConcentration at $0.5\%$, is already
reached by rounding alone. Using an identical budget definition across
both experiments allows results to be compared directly between the
industrial and UEA settings. We note that the smaller datasets, Libras
($180$ training instances) and EthanolConcentration ($261$ training
instances), still yield small absolute coreset sizes at the lowest
budgets (e.g., $1\%$ corresponds to $2$ and $3$ instances respectively), a
resolution limit worth keeping in mind when interpreting method
differences at the low-budget end of these two datasets.

\paragraph{Baselines.}

We compare against four of the baselines evaluated in the industrial
setting (Section~\ref{subsec:ind-setup}): Random, $k$-means, Herding, and Facility
location, all operating on the same truncated signature representation as
SOTS. In addition, we include $k$-medoids with Dynamic Time Warping,
solved via FasterPAM, as a baseline that instead operates directly on the
raw multivariate time series, without any signature encoding. Because
every other method in this comparison shares an identical signature-based
feature space, any performance gap between $k$-medoids with DTW and the
signature-based methods isolates the effect of the representation itself,
while differences among the signature-based methods isolate the effect of
the selection criterion alone.

\paragraph{SOTS Configuration.}

Since the raw channels of these datasets already form a regular,
synchronized multivariate path, no time-augmentation or transformation was used; signatures are computed directly from the raw
$d$-dimensional channels ($d=2$ for Libras and PenDigits, $d=3$ for
EthanolConcentration). Prior to signature computation, each instance is
independently $z$-normalized per channel along its own time axis,
$x \mapsto (x - \mu)/(\sigma + \epsilon)$ with $\epsilon = 10^{-8}$, using
that instance's own temporal mean and standard deviation. This
normalization prevents instances or channels with a wide range of raw
values, but comparatively small net displacement between their first and
last observations, from dominating the sliced Wasserstein distances used
for selection; without it, differences in raw scale rather than in
underlying shape could bias which instances are selected. Signatures are
truncated at level $N \in \{2, 3\}$ for Libras and PenDigits, and
$N \in \{2, 3, 4\}$ for EthanolConcentration, with the same per-level
rescaling scheme as the industrial setting (Section~\ref{subsec:ind-setup}) applied
to correct scale imbalance across retained levels. Because the resulting
signature dimensionality grows with both $d$ and $N$ (from $7$ at order
$2$ to $31$ at order $4$ for $d=2$; from $13$ to $121$ for $d=3$), we
scale the number of sliced Wasserstein projections accordingly to
maintain adequate coverage of the higher-dimensional space: $L = 100$ at
order $2$, $L = 200$ at order $3$, and $L = 500$ at order $4$. Sliced
Wasserstein distances are computed with $p=2$ using uniform projections
exclusively; the adaptive projection variants compared for the industrial
dataset~\ref{tab:results_projection} are not re-evaluated here.

\paragraph{Downstream Classifier.}

Unlike the industrial setting, where we adopt the Transformer architecture
of~\citep{anas}, coreset quality on the UEA datasets is
evaluated using ROCKET~\citep{rocket}, a random-convolutional-kernel
classifier shown to perform strongly and efficiently across the UEA
archive in the multivariate time series classification bake-off
of~\citep{uea2}. We use the \texttt{aeon} library's
\texttt{RocketClassifier}~\citep{aeon}, with default hyperparameters:
$10{,}000$ random convolutional kernels, whose extracted features are
standardized and passed to a \texttt{RidgeClassifierCV} classification
head, with the ridge regularization strength selected via the estimator's
built-in cross-validation. As with the industrial classifier, a new
ROCKET classifier is trained from scratch on each selected coreset and
evaluated once on the corresponding dataset's fixed test set. Coreset
selection and classifier training are entirely independent pipelines:
selection operates on the $z$-normalized signature representation
described above, while the classifier is trained directly on the raw,
un-normalized series as loaded via \texttt{aeon}, only the subset of
instances chosen by selection is shared between the two. We use
\texttt{aeon} for both the classifier and for loading the UEA datasets
themselves.

\paragraph{Evaluation Metrics.}

We report accuracy and macro-averaged F1, with F1-macro retained as our
primary metric for the same reason as in the industrial setting: it
weights every class equally regardless of its frequency. Unlike the
industrial dataset, no test-set balancing is required here, as the UEA
datasets are used with their canonical archive splits, which are already
close to class-balanced. AUPRC is not reported for this study, since it
was retained in the industrial setting specifically for comparability
with~\citep{anas}, whose classifier this experiment does not
reuse; ROCKET's default classification head
(\texttt{RidgeClassifierCV}) does not natively produce class probability
estimates, which AUPRC requires. For each (dataset, method, budget)
configuration, we report the mean and standard deviation across
$20$ runs.

\paragraph{Implementation and hardware.}
Coreset selection follows the same implementation as the industrial
setting (Section~\ref{subsec:ind-setup}): Python with PyTorch for the
GPU-parallelized components of the greedy selection procedures, and a
custom signature implementation. Downstream classification uses the
\texttt{aeon}~\citep{aeon} library (version $1.1.0$) for both dataset
loading and the ROCKET classifier. Unlike the
industrial experiments, this study is run on a different machine: a
single NVIDIA RTX 4000 Ada Generation GPU (20\,GB), using Python $3.9.25$, and PyTorch $2.8.0$ (CUDA $12.8$).

\subsubsection{Results}

\paragraph{Libras.}
Table~\ref{tab:sots_results_v3} reports results on Libras. Among the
three SOTS-Greedy truncation orders, order $2$ is the strongest for
most of the budget range, winning $10/14$ budgets on both accuracy and
F1-macro, against $0/14$ for order $3$; order $4$ wins the remaining
$3$ budgets, concentrated at the high-budget end ($50\%$, $75\%$,
$90\%$) rather than distributed across the curve. This is broadly
consistent with our hypothesis that when class-discriminative
information is itself geometric, as is the case for Libras'
hand-movement trajectories, low-order signature terms already capture
most of the relevant structure. SOTS-Greedy (order $2$) is also the strongest single method
overall against the full field of eight methods, winning $4/14$
budgets on accuracy and $4/14$ on F1-macro ($8/28$ total), narrowly
ahead of Facility location ($4/14$ accuracy, $3/14$ F1-macro; $7/28$
total) and $k$-medoids with DTW ($6/28$ total). All methods score
identically at the $0.5\%$ budget ($6.67\%$ accuracy, $0.83\%$
F1-macro), a mechanical artifact rather than a finding: at this budget
the coreset contains a single training instance regardless of
selection method. Herding performs poorly throughout, never winning a
single budget on either metric; notably, this weakness persists
despite Libras being class-balanced, unlike our industrial dataset,
indicating that class imbalance alone cannot explain Herding's failure
mode in general. We do not have a conclusive account of the mechanism
behind its poor performance on this dataset and leave it as an open
question.

\newcommand{\ms}[2]{\makecell{#1\\[-1pt]{\tiny\color{gray}$\pm$#2}}}
\newcommand{\msb}[2]{\makecell{\textbf{#1}\\[-1pt]{\tiny\color{gray}$\pm$#2}}}
\newcommand{\msu}[2]{\makecell{\underline{#1}\\[-1pt]{\tiny\color{gray}$\pm$#2}}}

\begin{table}[H]
\label{tab:libras}
\centering
\scriptsize
\setlength{\tabcolsep}{3pt}
\resizebox{\textwidth}{!}{%

\begin{tabular}{p{2.2cm}lcccccccccccccc}
\toprule
\textbf{Method} & \textbf{Metric} & \textbf{0.5\%} & \textbf{1\%} & \textbf{5\%} & \textbf{10\%} & \textbf{15\%} & \textbf{20\%} & \textbf{25\%} & \textbf{30\%} & \textbf{35\%} & \textbf{40\%} & \textbf{50\%} & \textbf{60\%} & \textbf{75\%} & \textbf{90\%}\\
\midrule

\multirow{2}{2.2cm}{\centering Random}
& Accuracy & \msb{6.67}{0.00} & \msu{9.83}{2.33} & \ms{34.64}{0.26} & \msu{49.11}{0.27} & \msu{57.61}{0.40} & \ms{64.50}{0.49} & \ms{70.11}{0.74} & \ms{70.56}{0.63} & \ms{72.25}{0.62} & \ms{75.69}{0.96} & \ms{83.61}{0.87} & \msu{87.25}{0.48} & \msb{90.94}{0.56} & \msb{91.11}{0.35} \\
& F1 Macro & \msb{0.83}{0.00} & \msu{2.86}{1.20} & \ms{22.09}{0.19} & \ms{37.16}{0.22} & \ms{50.35}{0.33} & \msu{61.75}{0.61} & \ms{68.46}{0.74} & \ms{69.86}{0.66} & \ms{71.75}{0.60} & \ms{75.40}{1.17} & \ms{83.21}{0.96} & \msu{87.27}{0.48} & \msb{90.95}{0.57} & \msb{91.14}{0.35} \\
\midrule

\multirow{2}{2.2cm}{\centering k-Medoids + DTW}
& Accuracy & \msb{6.67}{0.00} & \ms{8.86}{2.39} & \ms{30.31}{0.37} & \ms{47.50}{0.60} & \ms{55.14}{0.35} & \msu{65.33}{0.44} & \msb{75.92}{0.44} & \msb{76.50}{0.47} & \ms{76.78}{0.67} & \msb{80.06}{0.49} & \ms{83.03}{0.41} & \ms{84.14}{0.51} & \ms{88.25}{0.59} & \ms{89.86}{0.43} \\
& F1 Macro & \msb{0.83}{0.00} & \ms{2.15}{0.62} & \ms{19.81}{0.32} & \ms{42.15}{0.71} & \msu{51.00}{0.42} & \ms{61.68}{0.62} & \msb{74.03}{0.51} & \msb{74.63}{0.60} & \msu{74.72}{0.88} & \msb{78.52}{0.66} & \ms{82.78}{0.40} & \ms{84.04}{0.50} & \ms{88.33}{0.57} & \ms{89.83}{0.44} \\
\midrule

\multirow{2}{2.2cm}{\centering K-Means + Sig.}
& Accuracy & \msb{6.67}{0.00} & \ms{8.61}{2.20} & \msu{42.02}{0.44} & \ms{44.67}{0.87} & \ms{52.85}{0.19} & \ms{65.22}{0.71} & \ms{65.41}{0.59} & \msu{74.65}{0.55} & \msu{77.48}{0.42} & \msu{79.98}{0.60} & \msu{84.85}{0.76} & \ms{82.56}{0.57} & \ms{86.70}{0.67} & \ms{90.57}{0.66} \\
& F1 Macro & \msb{0.83}{0.00} & \ms{2.57}{1.03} & \msu{27.39}{0.33} & \ms{32.49}{0.63} & \ms{43.74}{0.28} & \ms{61.51}{0.78} & \ms{63.53}{0.59} & \ms{71.01}{0.80} & \ms{74.66}{0.56} & \msu{78.44}{0.66} & \msb{84.64}{0.76} & \ms{81.49}{0.72} & \ms{86.55}{0.72} & \ms{90.59}{0.65} \\
\midrule

\multirow{2}{2.2cm}{\centering Herding + Sig.}
& Accuracy & \msb{6.67}{0.00} & \ms{6.67}{0.00} & \ms{26.67}{0.00} & \ms{38.36}{0.37} & \ms{40.64}{0.36} & \ms{47.72}{0.35} & \ms{50.83}{0.67} & \ms{55.81}{0.76} & \ms{57.03}{0.71} & \ms{63.14}{0.56} & \ms{76.97}{0.51} & \ms{85.67}{0.45} & \ms{87.39}{0.36} & \ms{88.86}{0.45} \\
& F1 Macro & \msb{0.83}{0.00} & \ms{0.83}{0.00} & \ms{12.93}{0.04} & \ms{27.10}{0.36} & \ms{28.38}{0.27} & \ms{37.50}{0.32} & \ms{40.16}{0.75} & \ms{48.86}{0.70} & \ms{50.85}{0.87} & \ms{57.08}{0.74} & \ms{72.74}{0.51} & \ms{84.81}{0.56} & \ms{87.11}{0.36} & \ms{88.76}{0.44} \\
\midrule

\multirow{2}{2.2cm}{\centering Facility Loc. + Sig.}
& Accuracy & \msb{6.67}{0.00} & \ms{9.33}{2.51} & \msb{42.81}{0.28} & \ms{46.86}{0.54} & \ms{53.81}{0.79} & \ms{60.06}{0.30} & \ms{67.33}{0.38} & \ms{71.28}{0.61} & \msb{77.83}{0.46} & \ms{77.39}{0.61} & \msb{85.36}{0.40} & \msb{87.67}{0.65} & \msu{89.72}{0.48} & \ms{90.61}{0.72} \\
& F1 Macro & \msb{0.83}{0.00} & \ms{2.84}{1.03} & \msb{30.38}{0.18} & \ms{37.33}{0.46} & \ms{46.22}{0.76} & \ms{54.06}{0.34} & \ms{63.62}{0.51} & \ms{68.93}{0.63} & \msb{75.54}{0.67} & \ms{75.41}{1.09} & \msu{84.58}{0.50} & \msb{87.42}{0.67} & \ms{89.67}{0.47} & \ms{90.62}{0.71} \\
\midrule

\multirow{2}{2.2cm}{\centering SOTS-Greedy (ord.\ 2)}
& Accuracy & \msb{6.67}{0.00} & \msb{11.19}{2.02} & \ms{38.94}{0.17} & \msb{61.58}{0.62} & \msb{67.33}{0.76} & \msb{71.67}{0.46} & \msu{74.19}{0.64} & \ms{73.94}{0.46} & \ms{73.42}{0.59} & \ms{77.72}{0.39} & \ms{78.36}{0.12} & \ms{84.03}{0.65} & \ms{86.92}{0.51} & \ms{90.19}{0.51} \\
& F1 Macro & \msb{0.83}{0.00} & \msb{2.91}{0.59} & \ms{26.80}{0.16} & \msb{56.04}{0.61} & \msb{63.22}{0.66} & \msb{69.51}{0.53} & \msu{73.09}{0.67} & \msu{72.70}{0.45} & \ms{72.24}{0.59} & \ms{76.64}{0.46} & \ms{77.26}{0.12} & \ms{83.86}{0.66} & \ms{86.93}{0.52} & \ms{90.17}{0.51} \\
\midrule

\multirow{2}{2.2cm}{\centering SOTS-Greedy (ord.\ 3)}
& Accuracy & \msb{6.67}{0.00} & \ms{9.19}{4.07} & \ms{23.00}{0.71} & \ms{47.64}{0.72} & \ms{55.78}{0.41} & \ms{64.31}{0.49} & \ms{63.83}{0.43} & \ms{68.00}{0.48} & \ms{68.06}{0.28} & \ms{68.94}{0.39} & \ms{73.89}{0.66} & \ms{79.19}{0.45} & \ms{87.61}{0.53} & \ms{90.14}{0.30} \\
& F1 Macro & \msb{0.83}{0.00} & \ms{2.18}{1.10} & \ms{12.58}{0.46} & \ms{38.31}{0.62} & \ms{50.41}{0.42} & \ms{60.48}{0.46} & \ms{60.55}{0.52} & \ms{64.71}{0.57} & \ms{65.21}{0.27} & \ms{65.94}{0.34} & \ms{70.66}{0.93} & \ms{78.11}{0.44} & \ms{87.66}{0.55} & \ms{90.14}{0.29} \\
\midrule

\multirow{2}{2.2cm}{\centering \textbf{SOTS-Greedy (ord.\ 4)}}
& Accuracy & \msb{6.67}{0.00} & \msu{9.83}{2.58} & \ms{31.97}{0.28} & \ms{47.86}{0.20} & \ms{55.86}{0.54} & \ms{61.44}{0.44} & \ms{70.53}{0.76} & \ms{68.94}{0.78} & \ms{70.86}{0.60} & \ms{73.75}{0.49} & \ms{81.03}{0.66} & \ms{83.89}{0.72} & \ms{89.67}{1.00} & \msu{90.69}{0.35} \\
& F1 Macro & \msb{0.83}{0.00} & \ms{2.45}{0.80} & \ms{22.35}{0.19} & \msu{43.25}{0.18} & \ms{49.23}{0.70} & \ms{57.45}{0.65} & \ms{69.05}{0.78} & \ms{64.88}{0.77} & \ms{68.43}{0.70} & \ms{72.45}{0.55} & \ms{80.00}{0.85} & \ms{83.46}{0.79} & \msu{89.68}{1.00} & \msu{90.74}{0.35} \\

\bottomrule
\end{tabular}%
}
\caption{Accuracy and F1 Macro (\%) vs. coreset budget for "Libras" dataset, mean $\pm$ std. \textbf{Bold}: best per column; \underline{underline}: second-best per column.}
\label{tab:sots_results_v3}
\end{table}

\paragraph{PenDigits.}
Table~\ref{tab:sots_results} reports results on PenDigits. Unlike
Libras, the strongest single method overall is $k$-medoids with DTW,
our one baseline operating directly on raw sequences rather than
signatures, which wins $12/28$ cells ($43\%$) and dominates the
low-to-moderate budget range in particular, winning every cell from
$0.5\%$ through $20\%$ budget. This is not entirely unexpected: unlike
our industrial setting, where signals are irregularly and independently
sampled, motivating the choice of path signatures in the first place,
PenDigits consists of clean, regularly sampled, synchronized
trajectories, precisely the regime in which DTW-based methods are
well-suited and signature-based methods hold no inherent structural
advantage. Considered together, however, SOTS-Greedy remains
competitive even in this unfavorable regime: taking the better of
orders $2$ and $3$ at each budget, combined SOTS matches or exceeds the
single best non-SOTS method (including $k$-medoids with DTW) in
$14/28$ cells ($50\%$), and among signature-based methods specifically,
combined SOTS is the clear winner, taking $20/28$ cells ($71\%$)
against $k$-means, Herding, and Facility location (each $14\%$ or
below), all operating on the identical signature representation. Unlike
Libras, the two SOTS orders split budget ranges rather than one
dominating throughout: order $3$ is stronger at the lowest budgets
($0.5$–$5\%$) and again at high budgets ($50$–$75\%$), while order $2$
takes over in between ($30$–$40\%$ and $90\%$).

\begin{table}[h]
\centering
\scriptsize
\setlength{\tabcolsep}{3pt}
\resizebox{\textwidth}{!}{%
\begin{tabular}{p{2.2cm}lcccccccccccccc}
\toprule
\textbf{Method} & \textbf{Metric}
& \textbf{0.5\%} & \textbf{1\%} & \textbf{5\%}
& \textbf{10\%} & \textbf{15\%} & \textbf{20\%} & \textbf{25\%}
& \textbf{30\%} & \textbf{35\%} & \textbf{40\%} & \textbf{50\%}
& \textbf{60\%} & \textbf{75\%}
& \textbf{90\%}\\
\midrule

\multirow{2}{2.2cm}{\centering Random}
& Accuracy
& \ms{81.40}{0.20} & \ms{88.01}{0.22} & \ms{95.06}{0.21} & \ms{96.91}{0.11} & \ms{97.73}{0.12} & \ms{97.95}{0.13} & \ms{98.10}{0.16} & \msu{98.17}{0.13} & \ms{98.11}{0.14} & \ms{98.20}{0.12} & \msu{98.21}{0.14} & \ms{98.15}{0.15} & \ms{98.13}{0.13} & \msu{98.20}{0.15} \\
& F1 Macro
& \ms{80.77}{0.20} & \ms{87.61}{0.24} & \ms{95.12}{0.21} & \ms{96.95}{0.11} & \ms{97.75}{0.13} & \ms{97.97}{0.13} & \ms{98.13}{0.16} & \msu{98.19}{0.13} & \ms{98.14}{0.14} & \ms{98.22}{0.12} & \msu{98.23}{0.14} & \ms{98.17}{0.15} & \ms{98.15}{0.12} & \msu{98.22}{0.15} \\
\midrule

\multirow{2}{2.2cm}{\centering k-Medoids + DTW}
& Accuracy
& \msb{87.95}{0.13} & \msb{93.41}{0.17} & \msb{97.72}{0.09} & \msb{98.24}{0.13} & \msb{98.51}{0.13} & \msb{98.08}{0.18} & \msu{98.18}{0.15} & \ms{98.07}{0.17} & \ms{98.06}{0.12} & \ms{98.15}{0.12} & \ms{98.18}{0.14} & \ms{98.19}{0.12} & \ms{98.14}{0.13} & \ms{98.19}{0.12} \\
& F1 Macro
& \msb{87.48}{0.13} & \msb{93.52}{0.17} & \msb{97.74}{0.09} & \msb{98.25}{0.14} & \msb{98.52}{0.13} & \msb{98.09}{0.18} & \msu{98.19}{0.14} & \ms{98.08}{0.17} & \ms{98.07}{0.12} & \ms{98.16}{0.11} & \ms{98.19}{0.14} & \ms{98.20}{0.12} & \ms{98.16}{0.13} & \ms{98.21}{0.12} \\
\midrule

\multirow{2}{2.2cm}{\centering K-Means + Sig.}
& Accuracy
& \ms{82.68}{0.20} & \ms{87.95}{0.22} & \ms{96.50}{0.13} & \msu{97.47}{0.14} & \ms{97.39}{0.14} & \ms{97.90}{0.13} & \msb{98.34}{0.11} & \ms{97.91}{0.14} & \ms{98.11}{0.14} & \msu{98.27}{0.17} & \ms{98.16}{0.08} & \msu{98.20}{0.08} & \msu{98.22}{0.09} & \ms{98.15}{0.12} \\
& F1 Macro
& \ms{82.13}{0.21} & \ms{87.45}{0.22} & \ms{96.52}{0.13} & \msu{97.49}{0.14} & \ms{97.41}{0.14} & \ms{97.91}{0.12} & \msb{98.36}{0.11} & \ms{97.93}{0.14} & \ms{98.13}{0.14} & \msu{98.28}{0.17} & \ms{98.18}{0.08} & \msu{98.22}{0.08} & \msu{98.23}{0.09} & \ms{98.17}{0.12} \\
\midrule

\multirow{2}{2.2cm}{\centering Herding + Sig.}
& Accuracy
& \ms{41.53}{0.27} & \ms{45.08}{0.31} & \ms{70.12}{0.30} & \ms{75.82}{0.23} & \ms{77.07}{0.12} & \ms{78.08}{0.24} & \ms{83.65}{0.83} & \ms{87.40}{0.68} & \ms{92.68}{0.62} & \ms{94.43}{0.58} & \ms{96.36}{0.40} & \ms{97.64}{0.16} & \ms{98.03}{0.14} & \ms{98.12}{0.10} \\
& F1 Macro
& \ms{34.69}{0.28} & \ms{37.35}{0.31} & \ms{62.99}{0.30} & \ms{68.79}{0.27} & \ms{69.66}{0.13} & \ms{71.56}{0.46} & \ms{79.96}{0.98} & \ms{85.39}{0.90} & \ms{92.16}{0.78} & \ms{94.13}{0.69} & \ms{96.33}{0.41} & \ms{97.65}{0.16} & \ms{98.05}{0.15} & \ms{98.14}{0.10} \\
\midrule

\multirow{2}{2.2cm}{\centering Facility Loc. + Sig.}
& Accuracy
& \ms{80.85}{0.27} & \ms{87.73}{0.22} & \ms{95.52}{0.25} & \ms{97.44}{0.18} & \msu{97.80}{0.13} & \msu{97.97}{0.13} & \ms{97.93}{0.13} & \ms{98.09}{0.13} & \msu{98.18}{0.15} & \ms{98.13}{0.14} & \ms{98.21}{0.15} & \ms{98.19}{0.10} & \ms{98.15}{0.11} & \ms{98.16}{0.12} \\
& F1 Macro
& \ms{79.92}{0.31} & \ms{87.18}{0.22} & \ms{95.58}{0.25} & \ms{97.47}{0.18} & \msu{97.82}{0.13} & \msu{97.99}{0.12} & \ms{97.95}{0.13} & \ms{98.10}{0.13} & \msu{98.19}{0.15} & \ms{98.14}{0.14} & \ms{98.22}{0.15} & \ms{98.20}{0.10} & \ms{98.17}{0.11} & \ms{98.18}{0.12} \\
\midrule

\multirow{2}{2.2cm}{\centering SOTS-Greedy (ord.\ 2)}
& Accuracy
& \ms{81.32}{0.29} & \ms{88.71}{0.15} & \ms{94.22}{0.21} & \ms{97.22}{0.14} & \ms{97.66}{0.09} & \ms{97.73}{0.11} & \ms{97.94}{0.10} & \msb{98.28}{0.10} & \msb{98.36}{0.11} & \msb{98.38}{0.12} & \ms{98.20}{0.12} & \ms{98.17}{0.13} & \ms{98.15}{0.11} & \msb{98.27}{0.13} \\
& F1 Macro
& \ms{80.74}{0.29} & \ms{88.59}{0.14} & \ms{94.33}{0.21} & \ms{97.26}{0.14} & \ms{97.70}{0.09} & \ms{97.76}{0.11} & \ms{97.98}{0.10} & \msb{98.30}{0.10} & \msb{98.38}{0.11} & \msb{98.40}{0.12} & \ms{98.22}{0.11} & \ms{98.19}{0.13} & \ms{98.17}{0.11} & \msb{98.29}{0.13} \\
\midrule

\multirow{2}{2.2cm}{\centering \textbf{SOTS-Greedy (ord.\ 3)}}
& Accuracy
& \msu{84.05}{0.20} & \msu{90.17}{0.29} & \msu{96.56}{0.10} & \ms{97.26}{0.12} & \ms{97.37}{0.11} & \ms{97.67}{0.12} & \ms{97.87}{0.10} & \ms{98.02}{0.14} & \ms{98.16}{0.16} & \ms{98.24}{0.15} & \msb{98.25}{0.11} & \msb{98.30}{0.10} & \msb{98.31}{0.10} & \ms{98.19}{0.13} \\
& F1 Macro
& \msu{84.10}{0.20} & \msu{90.27}{0.29} & \msu{96.61}{0.10} & \ms{97.30}{0.12} & \ms{97.41}{0.11} & \ms{97.70}{0.12} & \ms{97.90}{0.10} & \ms{98.04}{0.14} & \ms{98.18}{0.16} & \ms{98.26}{0.15} & \msb{98.27}{0.11} & \msb{98.31}{0.10} & \msb{98.33}{0.10} & \ms{98.21}{0.13} \\

\bottomrule
\end{tabular}%
}
\caption{Accuracy and F1 Macro (\%) vs. coreset budget for "PenDigits" dataset, mean $\pm$ std over 20 runs. \textbf{Bold}: best per column; \underline{underline}: second-best per column.}
\label{tab:sots_results}
\end{table}

\paragraph{EthanolConcentration.}
Table~\ref{tab:sots_results_v1} reports results on EthanolConcentration,
included specifically as a dataset where class-discriminative
information is spectral rather than geometric. Facility location is the
clear overall winner, taking $10/28$ (budget, metric) cells outright
($36\%$), more than three times the win rate of $k$-medoids with DTW
($2/28$, $7\%$). Since Facility location operates on the identical
truncated signature representation as SOTS-Greedy, this indicates that
the signature representation itself is not the primary limitation on
this dataset, a signature-based selection method can still perform
well here, rather, it is specifically SOTS's sliced-Wasserstein
distributional-matching criterion that is mismatched to this setting.
Among the three SOTS truncation orders, order $3$ is decisively the
strongest, winning $11/14$ budgets on accuracy and $9/14$ on F1-macro,
against order $2$'s $1/14$ on each and order $4$'s $1/14$ and $3/14$
respectively; order $2$ never wins a single cell outright on this
dataset, a marked contrast to its dominance on Libras. Increasing
truncation further to order $4$ brings no additional benefit over order
$3$. This pattern is consistent with our overall hypothesis: on a
dataset where the informative signal is spectral rather than shape-based,
some information beyond the lowest-order signature terms is still
useful (order $3$ outperforms order $2$ substantially), but signature
information alone, regardless of order, remains a comparatively weak
proxy for the discriminative structure of this dataset, as the gap to
Facility location's overall win rate shows. As on Libras and PenDigits,
low-budget results at $0.5\%$ largely coincide across methods (a single
selected training instance), with Random's slightly different value
($24.71\%$ vs. $25.10\%$ accuracy) reflecting the dataset's four classes
not being perfectly balanced in the test set ($65$ vs. $66$ instances).\\

\begin{table}[H]
\centering
\scriptsize
\setlength{\tabcolsep}{3pt}
\resizebox{\textwidth}{!}{%

\begin{tabular}{p{2.2cm}lcccccccccccccc}
\toprule
\textbf{Method} & \textbf{Metric} & \textbf{0.5\%} & \textbf{1\%} & \textbf{5\%} & \textbf{10\%} & \textbf{15\%} & \textbf{20\%} & \textbf{25\%} & \textbf{30\%} & \textbf{35\%} & \textbf{40\%} & \textbf{50\%} & \textbf{60\%} & \textbf{75\%} & \textbf{90\%}\\
\midrule

\multirow{2}{2.2cm}{\centering Random}
& Accuracy & \msu{24.71}{0.00} & \ms{24.71}{0.00} & \msb{29.86}{0.60} & \ms{27.38}{0.77} & \ms{28.75}{0.71} & \ms{30.30}{0.65} & \ms{30.72}{0.70} & \ms{33.98}{1.01} & \ms{34.35}{0.84} & \ms{34.11}{0.79} & \msb{41.75}{1.18} & \msb{42.93}{0.99} & \msu{43.07}{0.96} & \ms{43.55}{1.19} \\
& F1 Macro & \msu{9.91}{0.00} & \ms{9.91}{0.00} & \msb{27.00}{0.61} & \ms{27.10}{0.80} & \ms{27.93}{0.66} & \ms{29.92}{0.65} & \ms{30.32}{0.67} & \ms{34.10}{1.01} & \ms{34.35}{0.81} & \ms{33.81}{0.81} & \msb{41.42}{1.14} & \msu{42.14}{0.96} & \msu{42.43}{0.85} & \msu{43.43}{1.15} \\
\midrule

\multirow{2}{2.2cm}{\centering k-Medoids + DTW}
& Accuracy & \msb{25.10}{0.00} & \ms{22.50}{0.50} & \ms{26.03}{0.39} & \ms{24.97}{0.79} & \ms{25.49}{0.88} & \ms{29.04}{0.76} & \ms{32.56}{0.79} & \ms{32.85}{0.85} & \ms{34.22}{0.94} & \ms{34.11}{0.86} & \ms{38.71}{0.88} & \ms{41.32}{0.85} & \msb{43.33}{1.08} & \ms{43.47}{0.74} \\
& F1 Macro & \msb{10.03}{0.00} & \ms{14.77}{0.40} & \ms{23.14}{0.47} & \ms{23.63}{0.86} & \ms{25.47}{0.92} & \ms{28.69}{0.73} & \ms{31.65}{0.79} & \ms{30.85}{0.80} & \ms{33.08}{0.94} & \ms{33.94}{0.86} & \ms{37.61}{0.86} & \ms{40.55}{0.84} & \msb{42.50}{1.16} & \ms{42.36}{0.76} \\
\midrule

\multirow{2}{2.2cm}{\centering K-Means + Sig.}
& Accuracy & \msb{25.10}{0.00} & \ms{21.44}{0.55} & \ms{23.50}{0.43} & \msb{31.33}{0.82} & \ms{28.00}{0.74} & \ms{29.92}{1.04} & \ms{34.56}{0.77} & \msu{37.88}{0.84} & \msu{37.98}{0.86} & \msu{39.92}{0.98} & \ms{40.06}{1.23} & \msu{42.08}{1.05} & \ms{41.50}{1.38} & \ms{41.80}{1.18} \\
& F1 Macro & \msb{10.03}{0.00} & \ms{15.53}{0.62} & \ms{19.82}{0.48} & \msb{29.69}{0.81} & \ms{26.11}{0.85} & \ms{29.67}{1.09} & \msu{33.67}{0.71} & \msu{36.97}{0.87} & \msu{37.90}{0.91} & \msb{40.36}{0.97} & \msu{40.79}{1.20} & \msb{42.17}{1.06} & \ms{41.79}{1.38} & \ms{41.66}{1.24} \\
\midrule

\multirow{2}{2.2cm}{\centering Herding + Sig.}
& Accuracy & \msb{25.10}{0.00} & \ms{24.23}{0.39} & \ms{25.65}{0.55} & \ms{24.50}{0.66} & \ms{25.77}{0.91} & \ms{28.54}{0.97} & \ms{32.02}{0.74} & \ms{32.04}{1.15} & \ms{35.32}{0.82} & \ms{35.78}{0.95} & \ms{37.79}{1.19} & \ms{36.76}{1.18} & \ms{39.49}{0.97} & \msb{44.55}{1.03} \\
& F1 Macro & \msb{10.03}{0.00} & \ms{14.83}{0.52} & \ms{18.25}{0.41} & \ms{23.30}{0.59} & \ms{25.46}{0.85} & \ms{27.27}{0.98} & \ms{30.10}{0.78} & \ms{31.19}{1.13} & \ms{34.67}{0.79} & \ms{34.23}{0.95} & \ms{36.00}{1.17} & \ms{36.23}{1.20} & \ms{39.86}{0.97} & \ms{43.39}{1.07} \\
\midrule

\multirow{2}{2.2cm}{\centering Facility Loc. + Sig.}
& Accuracy & \msb{25.10}{0.00} & \ms{23.14}{0.31} & \ms{24.64}{0.53} & \ms{26.25}{0.51} & \ms{27.16}{1.03} & \msb{32.85}{1.17} & \msb{35.87}{0.61} & \msb{39.42}{1.03} & \msb{40.32}{1.14} & \msb{40.43}{0.98} & \ms{40.03}{0.97} & \ms{40.96}{1.19} & \ms{41.93}{1.11} & \msu{43.65}{1.01} \\
& F1 Macro & \msb{10.03}{0.00} & \ms{11.31}{0.10} & \msu{24.36}{0.56} & \ms{25.35}{0.50} & \ms{26.91}{1.00} & \msb{32.71}{1.26} & \msb{35.71}{0.57} & \msb{39.13}{1.05} & \msb{39.95}{1.17} & \msu{40.07}{0.99} & \ms{40.32}{0.94} & \ms{41.17}{1.09} & \ms{41.12}{1.06} & \msb{43.86}{0.98} \\
\midrule

\multirow{2}{2.2cm}{\centering SOTS-Greedy (ord.\ 2)}
& Accuracy & \msb{25.10}{0.00} & \ms{23.27}{0.57} & \ms{24.18}{0.36} & \ms{28.11}{0.60} & \ms{28.82}{0.61} & \msu{32.50}{1.00} & \ms{31.65}{0.70} & \ms{32.18}{0.91} & \ms{31.91}{0.67} & \ms{33.81}{0.70} & \ms{37.16}{0.88} & \ms{38.72}{1.24} & \ms{40.71}{0.84} & \ms{42.79}{1.08} \\
& F1 Macro & \msb{10.03}{0.00} & \msu{18.00}{0.60} & \ms{21.51}{0.51} & \ms{27.15}{0.61} & \ms{27.97}{0.67} & \msu{32.41}{1.01} & \ms{30.47}{0.67} & \ms{32.40}{0.92} & \ms{31.96}{0.68} & \ms{33.92}{0.71} & \ms{36.70}{0.92} & \ms{38.33}{1.26} & \ms{40.41}{0.86} & \ms{42.72}{1.04} \\
\midrule

\multirow{2}{2.2cm}{\centering SOTS-Greedy (ord.\ 3)}
& Accuracy & \msb{25.10}{0.00} & \msb{25.17}{0.43} & \msu{27.72}{0.40} & \ms{26.83}{0.58} & \msb{31.00}{0.71} & \ms{32.31}{0.91} & \msu{34.58}{0.87} & \ms{36.02}{1.04} & \ms{35.73}{0.96} & \ms{36.46}{0.92} & \msu{40.30}{1.03} & \ms{41.66}{1.03} & \ms{42.76}{0.89} & \ms{43.54}{1.12} \\
& F1 Macro & \msb{10.03}{0.00} & \ms{16.71}{0.29} & \ms{23.12}{0.46} & \ms{25.07}{0.53} & \msu{29.01}{0.64} & \ms{30.91}{0.94} & \ms{33.32}{0.92} & \ms{34.75}{1.13} & \ms{35.07}{0.99} & \ms{35.90}{0.94} & \ms{38.86}{1.13} & \ms{39.35}{1.06} & \ms{42.31}{0.96} & \ms{42.86}{1.10} \\
\midrule

\multirow{2}{2.2cm}{\centering \textbf{SOTS-Greedy (ord.\ 4)}}
& Accuracy & \msb{25.10}{0.00} & \msu{24.98}{0.54} & \ms{24.45}{0.48} & \msu{28.63}{0.57} & \msu{30.30}{0.70} & \ms{32.38}{0.51} & \ms{32.50}{0.77} & \ms{33.37}{0.92} & \ms{33.21}{0.92} & \ms{34.01}{0.67} & \ms{38.00}{0.88} & \ms{38.59}{0.87} & \ms{39.53}{0.92} & \ms{43.19}{0.98} \\
& F1 Macro & \msb{10.03}{0.00} & \msb{18.21}{0.52} & \ms{22.98}{0.47} & \msu{28.36}{0.58} & \msb{29.67}{0.72} & \ms{31.57}{0.56} & \ms{31.89}{0.82} & \ms{33.39}{0.91} & \ms{33.78}{0.90} & \ms{34.93}{0.73} & \ms{38.85}{0.86} & \ms{38.75}{0.89} & \ms{39.66}{0.96} & \ms{42.64}{0.94} \\

\bottomrule
\end{tabular}%
}
\caption{Accuracy and F1 Macro (\%) vs. coreset budget for "EthanolConcentration" dataset, mean $\pm$ std over 30 runs. \textbf{Bold}: best per column; \underline{underline}: second-best per column.}
\label{tab:sots_results_v1}
\end{table}

Across the three UEA datasets, SOTS-Greedy does not emerge as a
dominant coreset selection method. Its clearest advantage appears on Libras, where
class-discriminative information is directly geometric and SOTS wins
outright against the baselines; on PenDigits and
EthanolConcentration, SOTS remains a strong and often the best
signature-based method, but is matched or exceeded overall by a
non-signature baseline ($k$-medoids with DTW) or a differently-motivated
signature-based selection criterion (Facility location).
Taken together, these results are best read not as a claim that SOTS
generalizes uniformly to multivariate time series classification, but
as evidence that its advantage is conditional on the structure of the
data: strongest where discriminative information is geometric, as
intended by its design. It is worth emphasizing, however, that SOTS was
motivated by and designed for irregularly sampled multivariate time
series, the setting of our industrial application, and the UEA archive
offers no readily available benchmark combining both geometric
discriminability and this form of irregularity; the datasets studied
here are all regularly sampled, testing generalization along the
geometric axis alone rather than the full regime SOTS targets. On the
question of truncation order, the best-performing order varies by
dataset, order $2$ on Libras, order $3$ on both PenDigits and
EthanolConcentration, with no single order dominating across all three;
we do not identify a universally preferable default, and selecting an
appropriate order for a new dataset likely requires the kind of
per-dataset comparison performed here. Finally, these results likely
understate SOTS's achievable performance on this benchmark: consistent
with our industrial ablation (Table~\ref{tab:results_projection}),
all UEA experiments use uniform sliced Wasserstein projections and a
plain, unaugmented signature construction, rather than the adaptive
projection variants or dataset-specific path augmentations that could
plausibly close some of the remaining gap to the strongest baselines on
PenDigits.

\newpage

\section{Discussion}

We introduced SOTS, a coreset selection method for irregular multivariate time series (IMTS) classification that combines truncated path signatures, representing each sample's cross-channel geometric structure exactly up to truncation order, with sliced Wasserstein optimal transport, used to greedily select a subset whose empirical distribution in signature space approximates that of the full dataset. Selection operates entirely on the signature representation, while downstream classifiers train independently on raw data, decoupling selection cost from the architecture ultimately deployed. We validated this approach on an industrial Instrumentation and Control (I\&C) false-positive discrepancy classification task, an irregularly sampled, class-imbalanced, real-world setting that directly motivated the method's design, and on three datasets from the UEA multivariate time series archive, testing generalization to regularly sampled data outside SOTS's original design regime. We further established a convergence guarantee for random coresets in signature space (Proposition~\ref{prop:convergence}), and showed empirically, via the coupling between sliced Wasserstein distance and downstream performance (Figure~\ref{fig:sw-perf}), that minimizing this distance is a meaningful proxy for downstream quality, particularly in the low-budget regime where coreset selection matters most in practice.

SOTS targets industrial and safety-critical domains where multivariate time series are irregularly and asynchronously sampled, and where classical methods, which typically assume regular sampling or rely on imputation, may perform poorly. Beyond I\&C testing, we expect similar conditions, independently sampled channels, validity or missingness indicators, class imbalance, to arise in other industrial settings. Our UEA generalization study also suggests a second class of applications: tasks where the discriminative signal is a trajectory's geometric shape even under regular sampling, such as gesture or motion classification, given SOTS's advantage on Libras and, among signature-based methods, on PenDigits. We present this as a direction suggested by our results rather than a validated claim, consistent with our finding that SOTS's advantage is conditional on data structure rather than universal. Applying it to a new dataset without first assessing whether its discriminative signal is geometric risks underperforming simpler alternatives. The choice of signature truncation order requires per-dataset tuning: order $2$ was preferable on Libras, order $3$ on PenDigits and EthanolConcentration, and we did not identify a rule for selecting it a priori. Our strongest results also depend on the sliced Wasserstein projection strategy: the adaptive variants that most improve performance (RPSW, IWRPSW) cost approximately $4.7\times$ the selection time of uniform projection (Table~\ref{tab:results_projection}), a trade-off we resolved in favor of uniform projection for our main results. Finally, our theoretical convergence guarantee currently covers random coresets only; it does not yet extend to the greedy selection procedure used in practice, leaving a gap between our theoretical and empirical contributions.

Several directions follow from these limitations. The clearest is computational: adaptive sliced Wasserstein projections, particularly IWRPSW, improve downstream performance over uniform projection but at a large computational cost; making this overhead practical, through a cheaper approximation to random-path direction sampling, would let future work choose more relevant points during the selection process. Our per-dataset comparison of truncation orders shows that order selection matters but does not, on its own, yield a rule for choosing it in advance; a principled, data-driven or theoretically motivated criterion for selecting truncation order given a new dataset is a natural next step. Our finding that decreasing sliced Wasserstein distance tracks a decreasing performance gap to the full-dataset ceiling, particularly at low budgets (Figure~\ref{fig:sw-perf}), suggests our greedy heuristic may be less efficient than a method that more directly approximates the SW-minimizing subset; designing such an algorithm, through better local search, a relaxation with provable approximation guarantees, or another optimization strategy, is a promising direction. Also, our convergence guarantee currently applies only to random coresets; extending it to the greedy coreset used in practice, or to other construction strategies, remains open, possibly through a suitably regularized sliced Wasserstein objective with more favorable theoretical properties, in the spirit of entropic regularization for standard optimal transport, though whether such a regularization preserves the properties needed for a greedy guarantee is itself an open question. More ambitiously, extending guarantees beyond distributional convergence to directly bound downstream model performance would be substantially more valuable, particularly in safety-critical, regulated domains such as the nuclear I\&C setting motivating this work, but is correspondingly harder, since it would need to account for the complexity and non-convexity of the downstream classifier itself. Finally, evaluating SOTS on irregularly sampled data from domains other than industrial I\&C testing would test generalization along the irregularity axis that our UEA study could not, and incorporating class-label information directly into selection may yield a new method for \textit{supervised} coreset selection.

\end{document}